\documentclass[10pt,twocolumn,letterpaper]{article}

\usepackage{iccv}
\usepackage{times}
\usepackage{epsfig}
\usepackage{graphicx}
\usepackage{amsmath}
\usepackage{amssymb}

\usepackage{refstyle}
\usepackage{bm} 
\usepackage{subcaption}
\usepackage{appendix}
\usepackage[font={small}]{caption}
\usepackage[abs]{overpic}

\usepackage[T1]{fontenc}
\usepackage[utf8]{inputenc}
\usepackage{authblk}

\usepackage{xcolor}
\usepackage{adjustbox}

\usepackage[breaklinks=true,bookmarks=false]{hyperref}

\iccvfinalcopy 

\def\iccvPaperID{948} 

\begin{document}

\title{
  SurfaceNet: An End-to-end 3D Neural Network for Multiview Stereopsis
}


\author[1]{Mengqi Ji\thanks{mji@connect.ust.hk}}
\author[3]{Juergen Gall}
\author[2]{Haitian Zheng}
\author[2]{Yebin Liu}
\author[2]{Lu Fang\thanks{luvision.net@gmail.com}}  
\affil[1]{Hong Kong University of Science and Technology, Hong Kong, China}
\affil[2]{Tsinghua University, Beijing, China}
\affil[3]{University of Bonn, Bonn, Germany}

\renewcommand\Authands{ and }

\maketitle

\begin{abstract}
This paper proposes an end-to-end learning framework for multiview stereopsis. We term the network \textit{SurfaceNet}. 
    It takes a set of images and their corresponding camera parameters as input
    and directly infers the 3D model.
The key advantage of the framework is that both photo-consistency
    as well geometric relations of the surface structure can be directly learned for the purpose of multiview stereopsis in an end-to-end fashion.
    SurfaceNet is a fully 3D convolutional network which is achieved by encoding the camera parameters together with the images in a 3D voxel representation.    
    We evaluate SurfaceNet on the large-scale DTU benchmark. 
\end{abstract}

%
\section{Introduction}

In multiview stereopsis (MVS), a dense model of a 3D object is reconstructed from a set of images with known camera parameters. 
This classic computer vision problem has been extensively studied and the standard pipeline involves a number of separate steps~\cite{campbell2008using,tola2012efficient}.
In available multiview stereo pipelines, sparse features are first detected and then propagated to a dense point cloud for covering the whole surface \cite{furu2010accurate,HabbeckeK07}, or multiview depth maps are first computed followed with a depth map fusion step to obtain the 3D reconstruction of the object \cite{LiuCDX09,MerrellAWMFYNP07}. A great variety of approaches have been proposed to improve different steps in the standard pipeline. For instance, the works~\cite{tola2012efficient, campbell2008using, galliani2015massively} have focused on improving the depth map generation using MRF optimization, photo-consistency enforcement, or other depth map post-processing operations like denoising or interpolation. Other approaches have focused on more advanced depth fusion algorithms \cite{jancosek2011multi}. 
\par

Recent advances in deep learning have been only partially integrated. For instance, \cite{zbontar2015computing} uses a CNN instead of hand-crafted features for finding correspondences among image pairs and \cite{7780960} predicts normals for depth maps using a CNN, which improves the depth map fusion. The full potential of deep learning for multiview stereopsis, however, can only be explored if the entire pipeline is replaced by an end-to-end learning framework that takes the images with camera parameters as input and infers the surface of the 3D object. 
Apparently, such end-to-end scheme has the advantage that photo-consistency and geometric context for dense reconstruction can be directly learned from data without the need of manually engineering separate processing steps.                 

\begin{figure}[t]
    \centering
    \newcommand{\colw}{0.3}
    \newcommand{\figw}{1.1} 
    \begin{subfigure}[t]{\colw\linewidth}
        \includegraphics[width=\figw\textwidth,trim={2cm 0cm 0cm 0cm},clip]{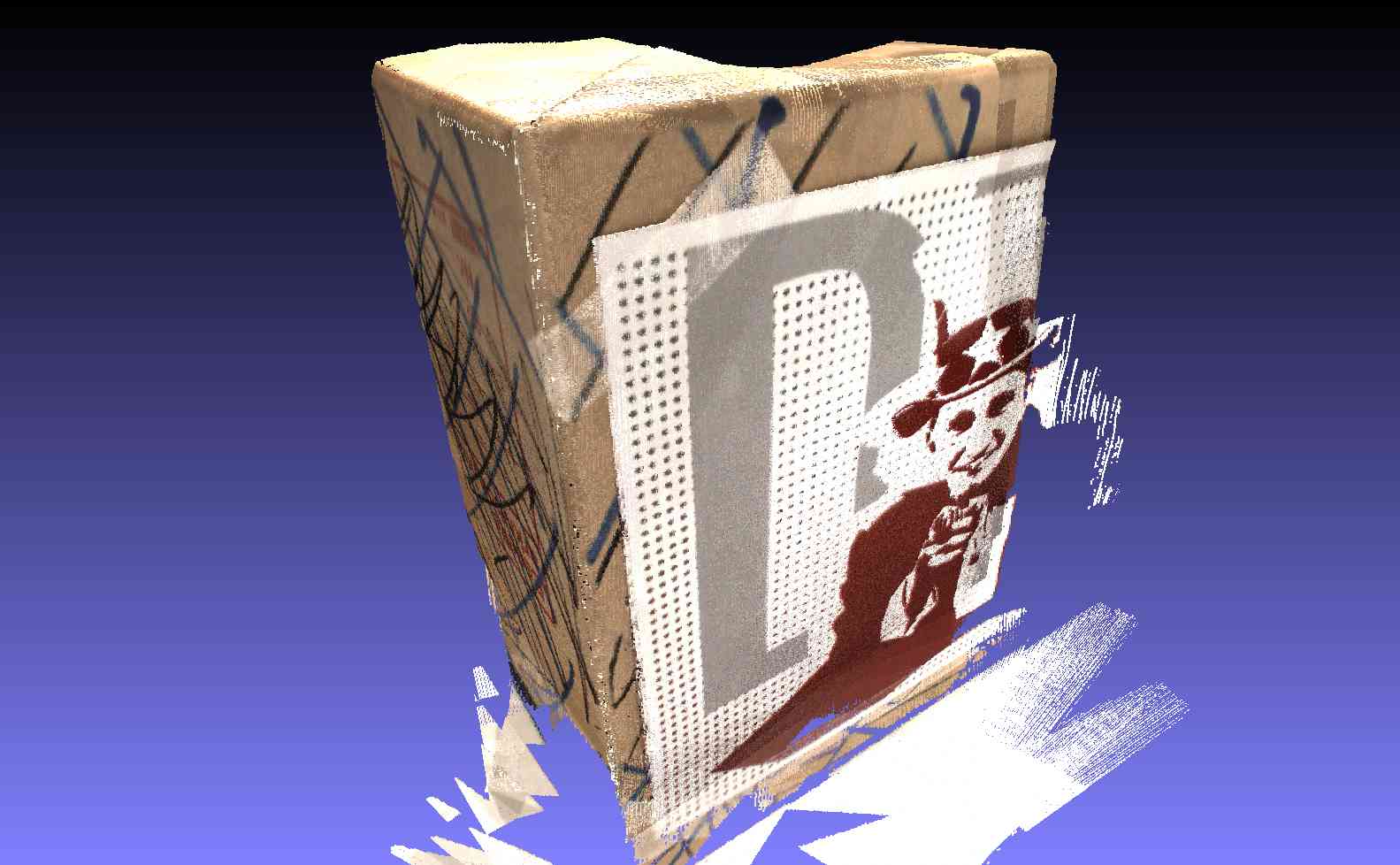}
        \caption{reference model}
    \end{subfigure}%
    ~ 
    \begin{subfigure}[t]{\colw\linewidth}
        \includegraphics[width=\figw\textwidth,trim={2cm 0cm 0cm 0cm},clip]{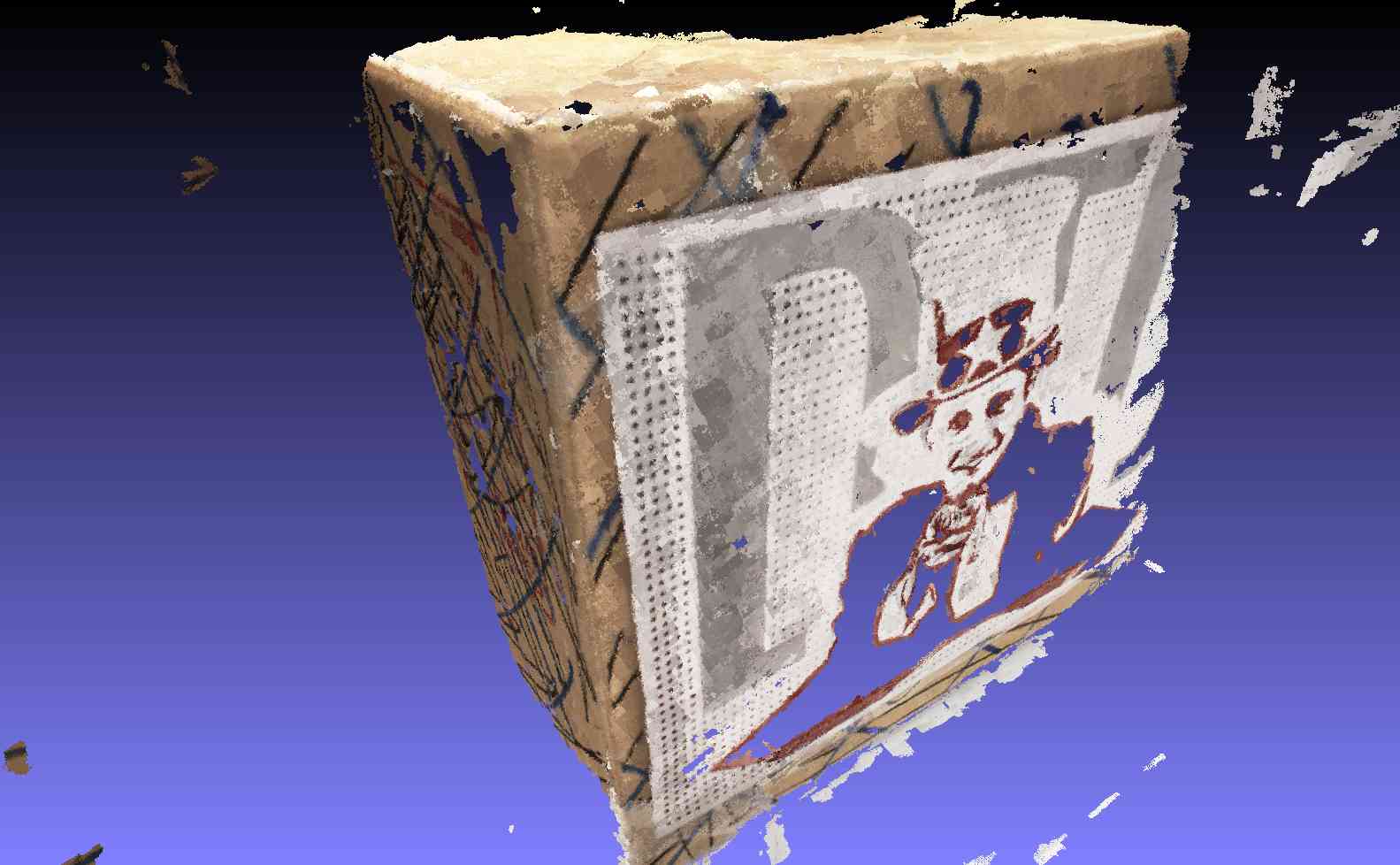}
        \caption{\textbf{SurfaceNet}}
    \end{subfigure}
    ~ 
    \begin{subfigure}[t]{\colw\linewidth}
        \includegraphics[width=\figw\textwidth,trim={2cm 0cm 0cm 0cm},clip]{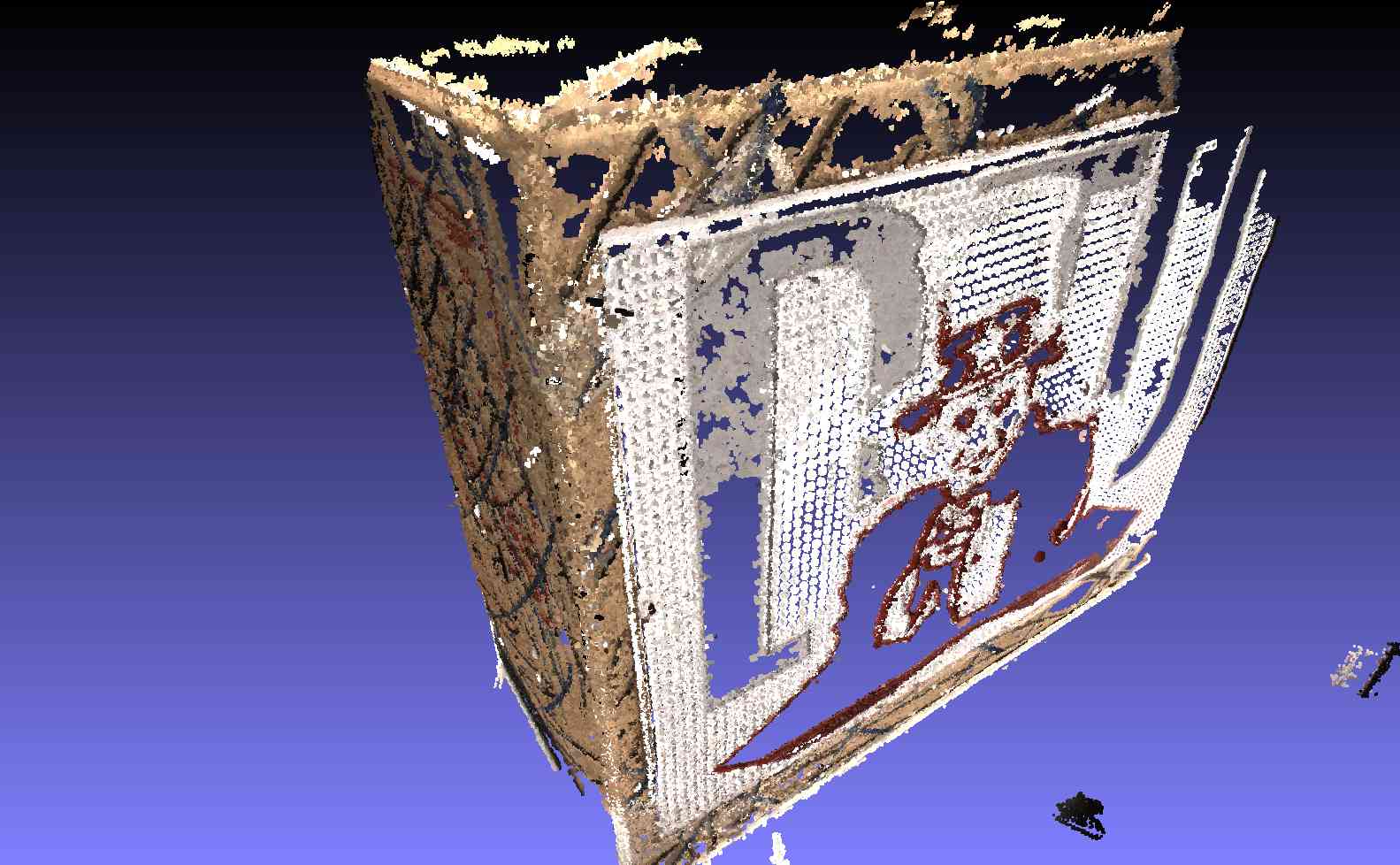}
        \caption{\textit{camp} \cite{campbell2008using}}
    \end{subfigure}
    ~ 
    \begin{subfigure}[t]{\colw\linewidth}
        \includegraphics[width=\figw\textwidth,trim={2cm 0cm 0cm 0cm},clip]{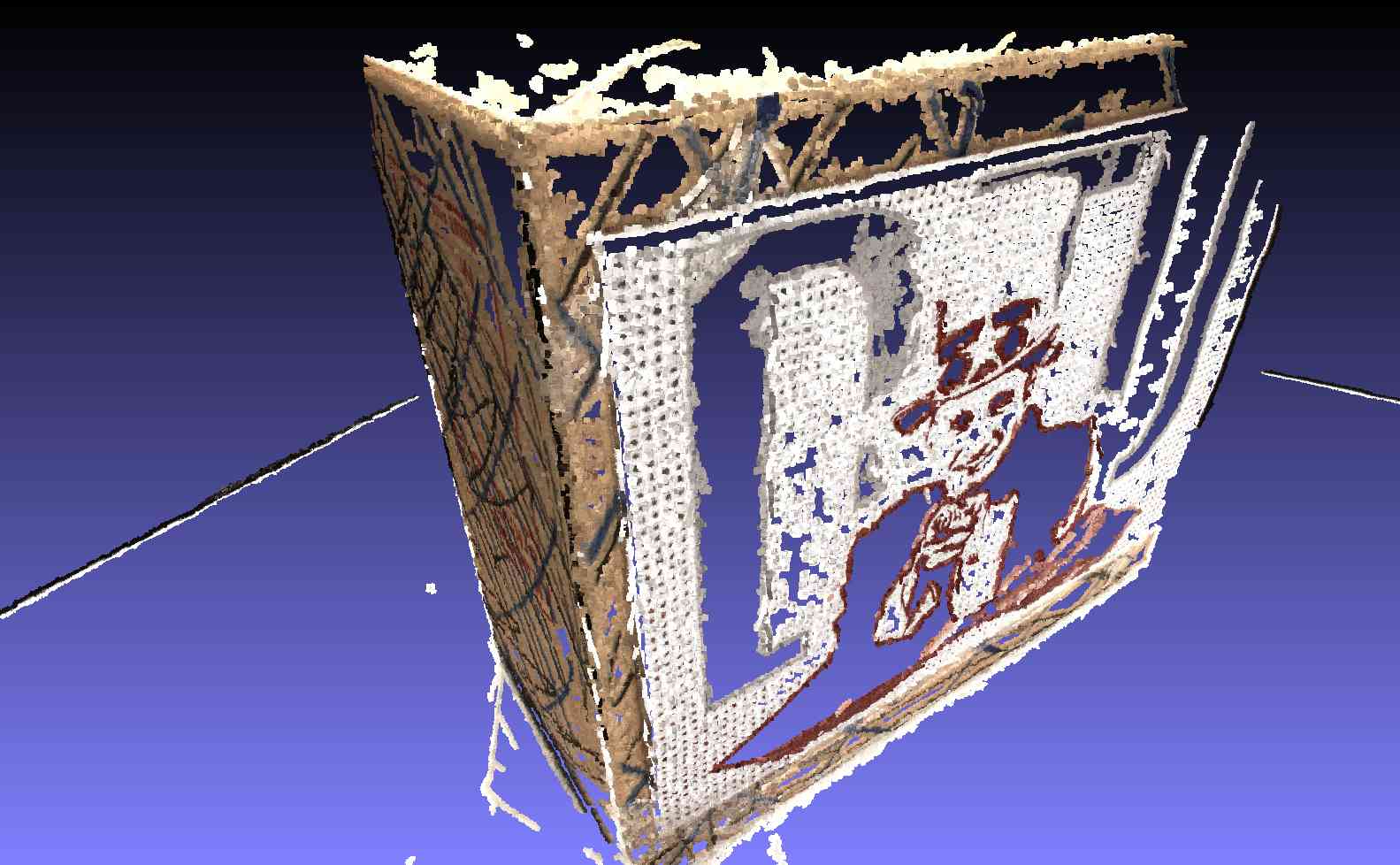}
        \caption{\textit{furu} \cite{furu2010accurate}}
    \end{subfigure}
    ~ 
    \begin{subfigure}[t]{\colw\linewidth}
        \includegraphics[width=\figw\textwidth,trim={2cm 0cm 0cm 0cm},clip]{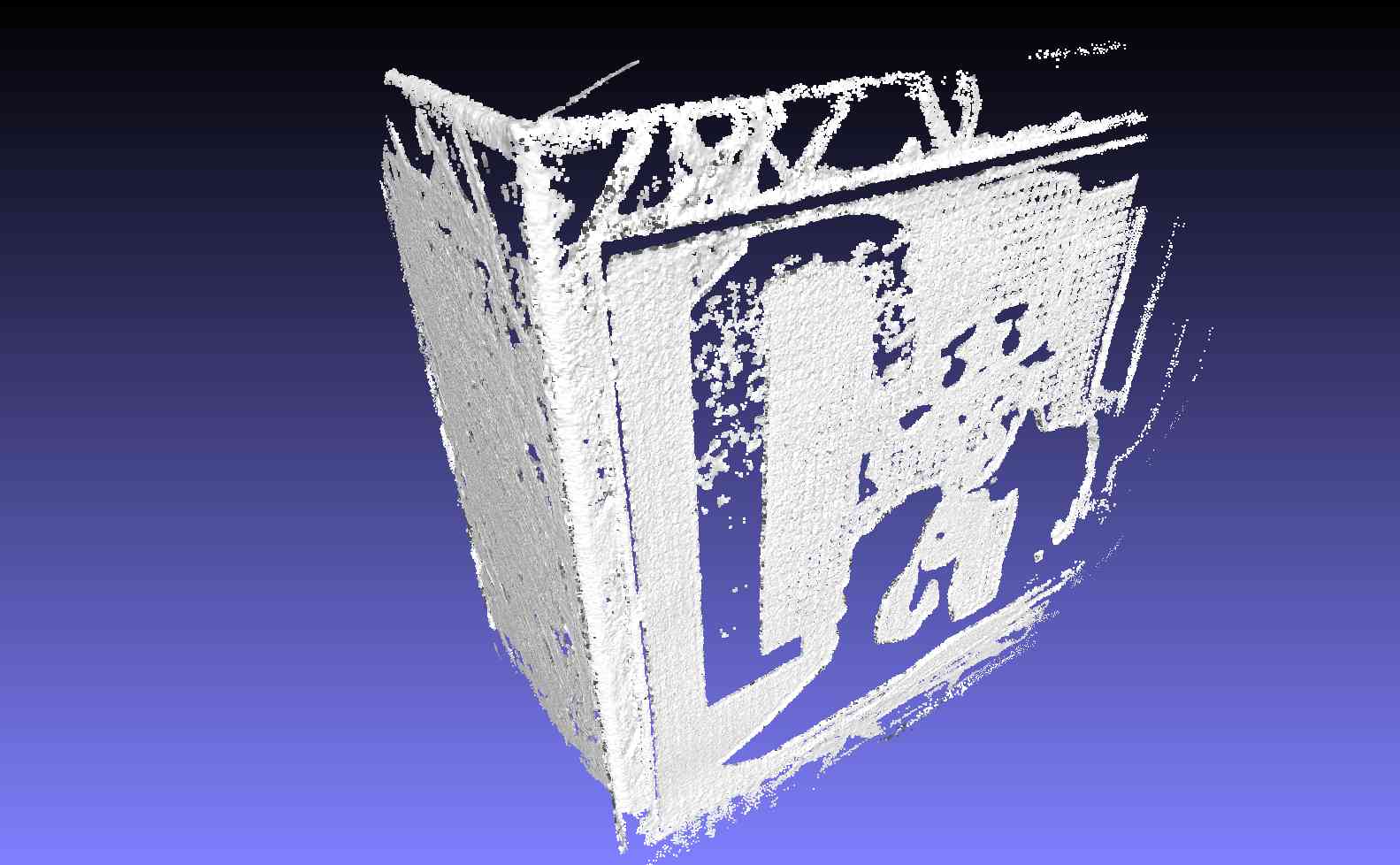}
        \caption{\textit{tola} \cite{tola2012efficient}}
    \end{subfigure}
    ~ 
    \begin{subfigure}[t]{\colw\linewidth}
        \includegraphics[width=\figw\textwidth,trim={2cm 0cm 0cm 0cm},clip]{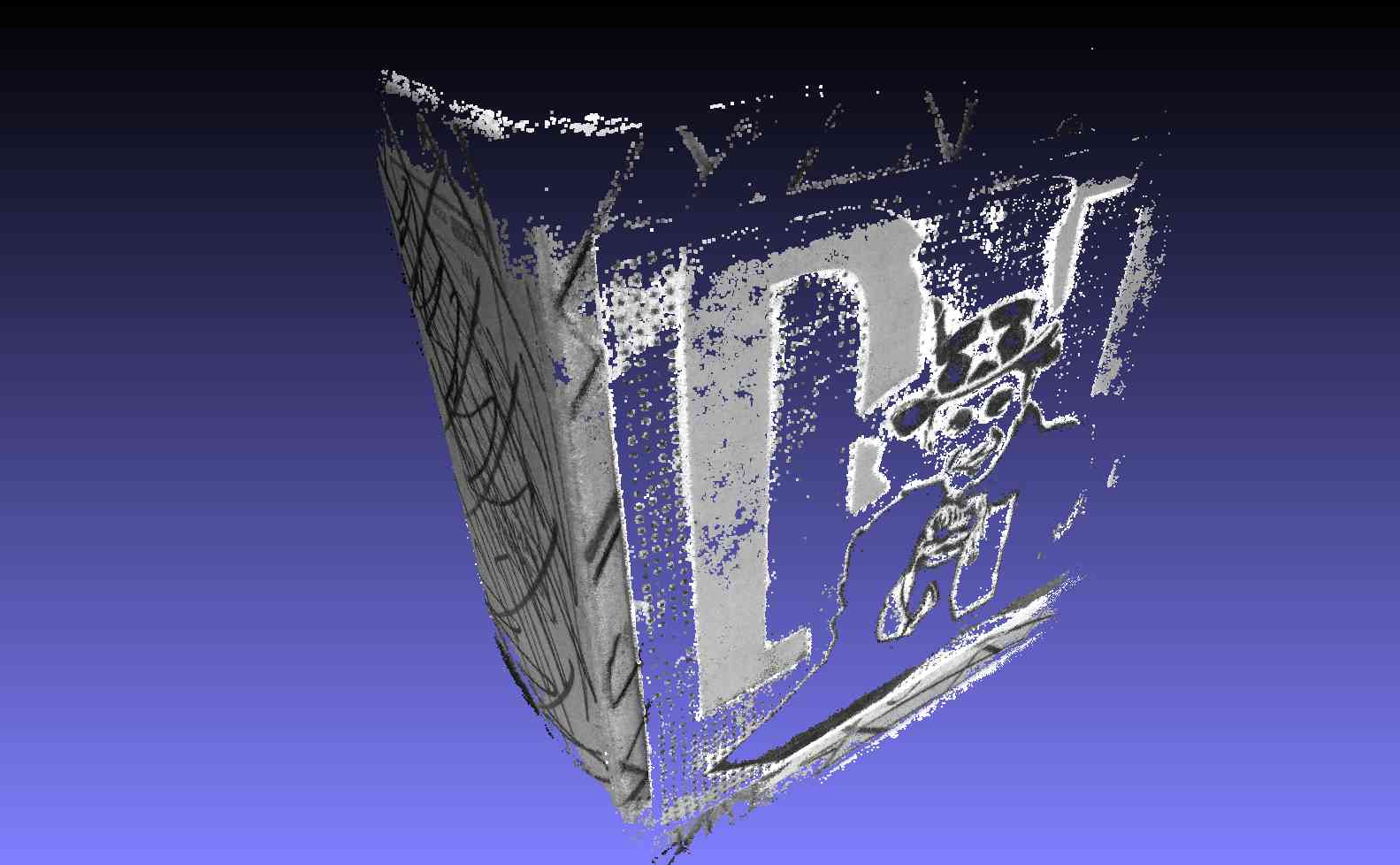}
        \caption{\textit{Gipuma} \cite{galliani2015massively}}
    \end{subfigure}%
    \caption{Reconstruction of the model $13$ of the DTU dataset \cite{aanaes2016large} in comparison to \cite{campbell2008using,furu2010accurate,tola2012efficient,galliani2015massively}.
     Our end-to-end learning framework provides a relatively complete reconstruction.}
   \label{fig:showcase}
\end{figure}

Instead of improving the individual steps in the pipeline, we therefore
propose the first end-to-end learning framework for multiview stereopsis, named \textit{SurfaceNet}. Specifically, SurfaceNet is a 3D
convolutional neural network that can process two or more
views and the loss function is directly computed based on the predicted
surface from all available views. In order to obtain a fully
convolutional network, a novel representation for each available
viewpoint, named \textit{colored voxel cube} (CVC), is proposed to implicitly encode the camera parameters via a straightforward perspective projection operation outside the network. 
Since the network predicts surface probabilities, we obtain a reconstructed surface by an additional binarization step. 
In addition, an optional thinning operation can be
applied to reduce the thickness of the surface. Besides of binarization
and thinning, SurfaceNet does not require any additional post-processing
or depth fusion to obtain an accurate
and complete reconstruction.

\section{Related Works}

Works in the multiview stereopsis (MVS) field can be roughly categorised into volumetric methods and depth maps fusion algorithms. While earlier works like space carving \cite{spacecarving99,Seitz1999} mainly use a volumetric representation, current state-of-the-art MVS methods focus on depth map fusion algorithms~\cite{tola2012efficient,campbell2008using, galliani2015massively}, 
    which have been shown to be more competitive in handling large datasets in practice. 
As our method is more related to the second category, our survey mainly covers depth map fusion algorithms. A more comprehensive overview of MVS approaches is given in the tutorial article~\cite{furukawa2015multi}.
\par

The depth map fusion algorithms first recover depth maps \cite{xu2013novel} from view pairs by matching similarity patches \cite{barnes2009patchmatch,pang2014self,zheng2015motion} along epipolar line and then fuse the depth maps to obtain a 3D reconstruction of the object \cite{tola2012efficient,campbell2008using, galliani2015massively}. 
In order to improve the fusion accuracy, \cite{campbell2008using} mainly learns several sources of the depth map outliers.
\cite{tola2012efficient} is designed for ultra high-resolution image sets and uses a robust decriptor for efficient matching purposes.
The \textit{Gipuma} algorithm proposed in \cite{galliani2015massively} is a massively parallel method for multiview matching built on the idea of patchmatch stereo \cite{bleyer2011patchmatch}. 
Aggregating image similarity across multiple views, \cite{galliani2015massively} can obtain more accurate depth maps.
The depth fusion methods usually contain several manually engineered steps, such as point matching, depth map denoising, and view pair selection.
Compared with the mentioned depth fusion methods, the proposed SurfaceNet infers the 3D surface 
    with thin structure directly from multiview images without the need of manually engineering separate processing steps.
\par

The proposed method in \cite{furu2010accurate} describes a patch model that consists of a quasi-dense set of rectangular patches covering the surface. 
The approach starts from a sparse set of matched keypoints that are repeatedly expanded to nearby pixel correspondences before filtering the false matches using visibility constraints.
However, the reconstructed rectangular patches cannot contain enough surface detail, which results in small holes around the curved model surface.
In contrast, our data-driven method predicts the surface with fine geometric detail and has less holes around the curved surface.
\par

Convolutional neural networks have been also used in the context of MVS. In \cite{7780960}, a CNN is trained to predict the normals of a given depth map based on image appearance. The estimated normals are then used to improve the depth map fusion. Compared to its previous work \cite{galliani2015massively}, the approach increases the completeness of the reconstruction at the cost of a slight decrease in accuracy. 
Deep learning has also been successfully applied to other 3D applications like volumetric shape retrieval or object classification \cite{wu20153d,song2015deep,maturana2015voxnet,qi2016volumetric}. In order to simplify the retrieval and representation of 3D shapes by CNNs, \cite{sinha2016deep} introduces a representation, termed geometry image, which is a 2D representation that approximates a 3D shape. Using the 2D representation, standard 2D CNN architectures can be applied.
While the 3D reconstruction of an object from a single or multiple views using convolutional neural networks has been studied in~\cite{tatarchenko2016multi,choy20163d}, the approaches focus on a very coarse reconstruction without geometric details when only one or very few images are available. In other words, these methods are not suitable for MVS.
The proposed SurfaceNet is able reconstruct large 3D surface models with detailed surface geometry.
\par

\section{Overview}

We propose an end-to-end learning framework that takes a set of images and their corresponding camera parameters as input and infers the 3D model.
To this end, we voxelize the solution space and propose a convolutional neural network (CNN) that predicts for each voxel $x$ a binary attribute $s_x \in \{0,1\}$ depending on whether the voxel is on the surface or not. We call the network, which reconstructs a 2D surface from a 3D voxel space, \textit{SurfaceNet}. It can be considered as an analogy to object boundary detection \cite{xie2015holistically}, which predicts a 1D boundary from 2D image input. 

We introduce SurfaceNet in Section~\ref{intraVolume} first for the case of two views and generalize the concept to multiple views in Section~\ref{multiViewFusion}. In Section~\ref{implementation} we discuss additional implementation details like the early rejection of empty volumes that speed up the computation.

\section{SurfaceNet}\label{intraVolume}

Given two images $I_i$ and $I_j$ for two views $v_i$ and $v_j$ of a scene with known camera parameters and a voxelization of the scene denoted by a 3D tensor $C$, 
our goal is to reconstruct the 2D surface in $C$ by estimating for each voxel $x\in C$ if it is on the surface or not, \ie $s_x \in \{0,1\}$.

To this end, we propose an end-to-end learning framework, which automatically learns both photo-consistency and geometric relations of the surface structure. Intuitively, for accurate reconstruction, the network requires the images $I_i$ and $I_j$ as well as the camera parameters. However, direct usage of $I_i$, $I_j$ and their corresponding camera parameters as input would unnecessarily increase the complexity of the network since such network needs to
learn the relation between the camera parameters and the projection of a voxel onto the image plane in addition to the reconstruction. Instead, we propose a 3D voxel representation that encodes the camera parameters implicitly such that our network can be fully convolutional.


\begin{figure}[h] 
\centering
    \includegraphics[width=0.8\linewidth]{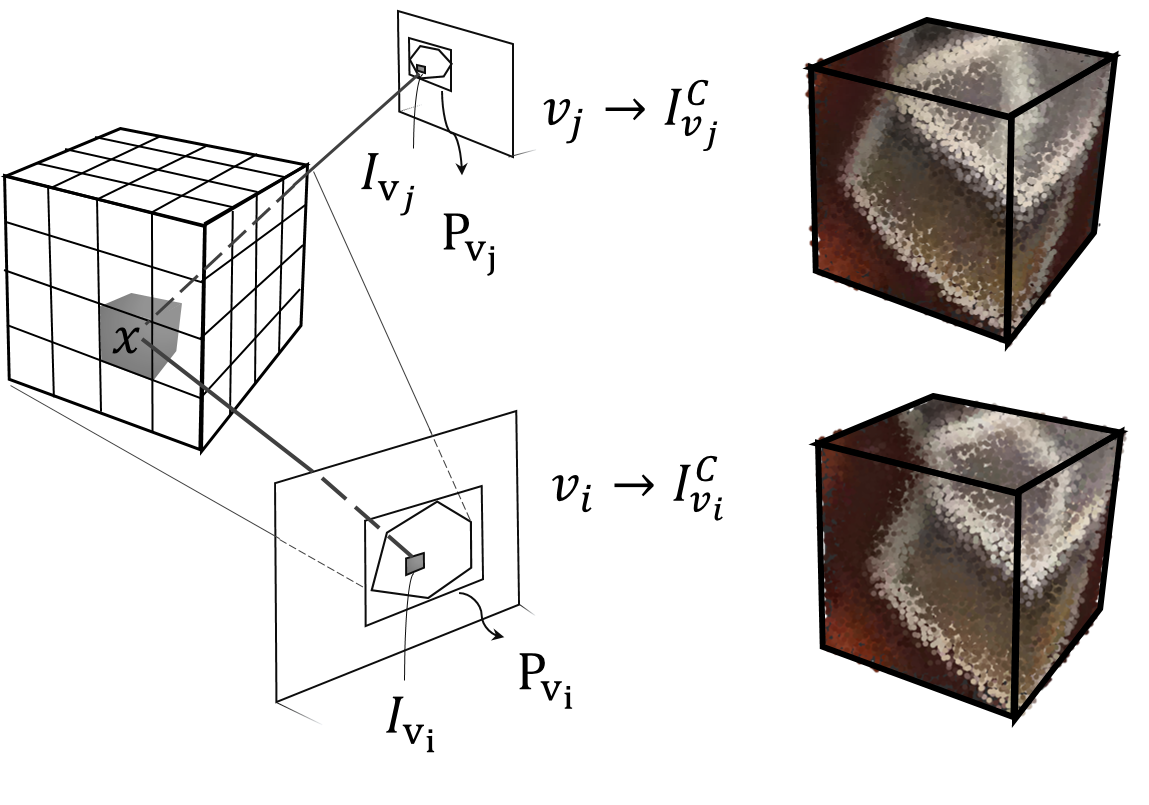} 
    \caption{Illustration of two Colored Voxel Cubes (CVC).}
    \label{fig:coloredCube}
\end{figure}

We denote our representation as \textit{colored voxel cube} (CVC), which is computed for each view and illustrated in Fig.~\ref{fig:coloredCube}. For a given view $v$, we convert the image $I_v$ into a 3D colored cube $I^C_v$ by projecting each voxel $x \in C$ onto the image $I_v$ and storing the RGB values $i_x$ for each voxel respectively. For the color values, we subtract the mean color~\cite{simonyan2014very}.
Since this representation is computed for all voxels $x \in C$, the voxels that are on the same projection ray have the same color $i_x$. In other words, the camera parameters are encoded with CVC. As a result, we obtain for each view a projection-specific stripe pattern as illustrated in Fig.~\ref{fig:coloredCube}.

%
%

\subsection{SurfaceNet Architecture}

%
%

\begin{figure}[htbp] 
\centering
\includegraphics[width=\linewidth]{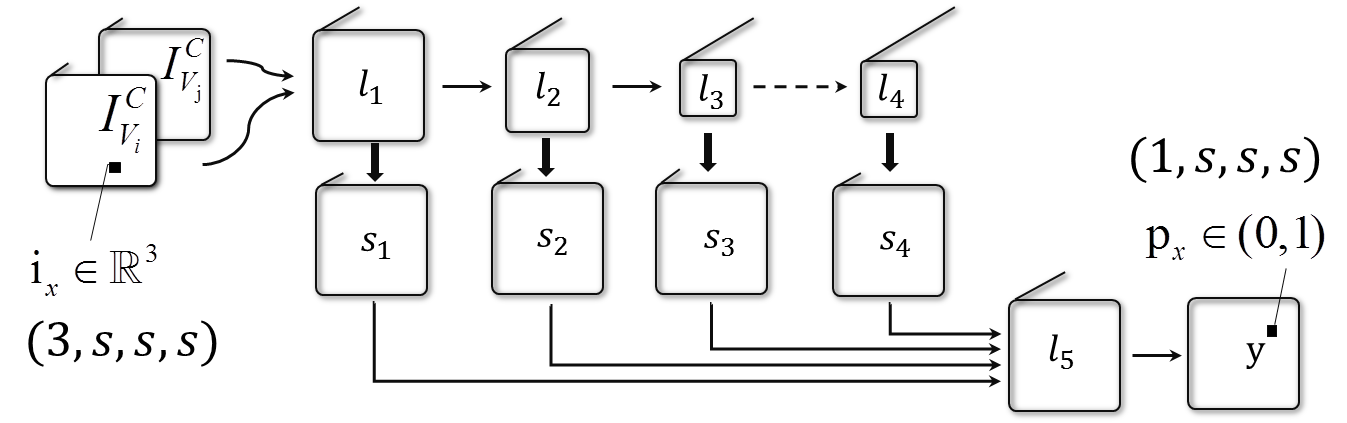} 
\caption{SurfaceNet takes two CVCs from different viewpoints as input.
    Each of the RGB-CVC is a tensor of size $(3,s,s,s)$.
    In the forward path, there are four groups of convolutional layers. The $l_{4.\cdot}$ layers are 2-dilated convolution layers.
    The side layers $s_{i}$ extract multi-scale information, which are aggregated into the output layer $y$
        that predicts the on-surface probability for each voxel position. The output has the size $(1,s,s,s)$.}
\label{fig:surfaceNet}
\end{figure}
The architecture of the proposed SurfaceNet is shown in Fig.~\ref{fig:surfaceNet}. It takes two colored voxel cubes from two different viewpoints as input and predicts for each voxel $x \in C$ the confidence $p_x \in (0,1)$, which indicates if a voxel is on the surface. While the conversion of the confidences into a surface is discussed in Section~\ref{inference}, we first describe the network architecture.

The detailed network configuration is summarized in Table~\ref{table:3DCNN-tab}. The network input is a pair of colored voxel cubes, where each voxel stores three RGB color values. For cubes with $s^3$ voxels, the input is a tensor of size $6\times s\times s\times s$. The basic building blocks of our model are 3D convolutional layers $l(\cdot)$, 3D pooling layers $p(\cdot)$ and 3D up-convolutional layers $s(\cdot)$,
where $l_{n.k}$ represents the $k$th layer in the $n$th group of convolutional layers. Additionally, a rectified linear unit (ReLU) is appended to each convolutional layer $l_i$ and the sigmoid function is applied to the layers $s_i$ and $y$. In order to decrease the training time and increase the robustness of training, batch normalization \cite{ioffe2015batch} is utilized in front of each layer.
The layers in $l_4$ are dilated convolutions \cite{yu2015multi} with dilation factor of 2.
They are designed to exponentially increase the receptive field without loss of feature map resolution.
The layer $l_{5.k}$ increases the performance by aggregating multi-scale contextual information from the side layers $s_i$
    to consider multi-scale geometric features.
Since the network is fully convolutional, the size of the CVC cubes can be adaptive. The output is always the same size as the CVC cubes.

\begin{table}[htbp]
    \begin{tabular}{cccc} 
    \hline
    layer name & type & output size & kernel size \\
    \hline
    input                      & CVC         & $(6, s, s, s)$    & - \\
    $l_{1.1},l_{1.2},l_{1.3}$   & conv      & $(32, s, s, s)$    & $(3,3,3)$ \\
    $s_1$                       & upconv      & $(16, s, s, s)$    & $(1,1,1)$   \\
    $p_1$                       & pooling   & $(32, \frac{s}{2}, \frac{s}{2}, \frac{s}{2})$  & $(2,2,2)$  \\
    $l_{2.1},l_{2.2},l_{2.3}$   & conv      & $(80, \frac{s}{2}, \frac{s}{2}, \frac{s}{2})$    & $(3,3,3)$ \\
    $s_2$                       & upconv    & $(16, s, s, s)$    & $(1,1,1)$   \\
    $p_2$                       & pooling   & $(80, \frac{s}{4}, \frac{s}{4}, \frac{s}{4})$  & $(2,2,2)$  \\
    $l_{3.1},l_{3.2},l_{3.3}$   & conv      & $(160, \frac{s}{4}, \frac{s}{4}, \frac{s}{4})$    & $(3,3,3)$ \\
    $s_3$                       & upconv    & $(16, s, s, s)$    & $(1,1,1)$   \\
    $l_{4.1},l_{4.2},l_{4.3}$   & dilconv      & $(300, \frac{s}{4}, \frac{s}{4}, \frac{s}{4})$    & $(3,3,3)$ \\
    $s_4$                       & upconv    & $(16, s, s, s)$    & $(1,1,1)$   \\
    $l_{5.1},l_{5.2}$           & conv      & $(100, s, s, s)$    & $(3,3,3)$ \\
    $y$                       & conv    & $(1, s, s, s)$    & $(1,1,1)$   \\
    \hline
    \end{tabular}
\captionof{table}{Architecture of SurfaceNet. A rectified linear activation function is used after each convolutional layer except $y$,
       and a sigmoid activation function is used after the up-convolutional layers and the output layer to normalize the output.
} \label{table:3DCNN-tab}
\end{table}

\subsection{Training}
As the SurfaceNet is a dense prediction network, \ie, the network predicts the surface confidence for each voxel, we compare the prediction per voxel $p_x$ with the ground-truth $\hat{s}_x$. For training, we use a subset of the scenes from the DTU dataset \cite{aanaes2016large} which provides images, camera parameters, and reference reconstructions obtained by a structured light system. A single training sample consists of a cube $\hat{S}^C$ cropped from a 3D model and two CVC cubes $I^C_{v_i}$ and $I^C_{v_j}$ from two randomly selected views $v_i$ and $v_j$.
Since most of the voxels do not contain the surface, \ie $\hat{s}_x=0$, we weight the surface voxels by
\begin{equation}
\alpha = \frac{1}{|\mathcal{C}|}\sum_{C \in \mathcal{C}}\frac{\sum_{x \in C}(1-\hat{s}_x)}{|C|}, 
\end{equation}
where $\mathcal{C}$ denotes the set of sampled training samples, and the non-surface voxels by $1-\alpha$. We use a class-balanced cross-entropy function as loss for training, \ie for a single training sample $C$ we have:
\begin{align} \label{eq:balancedEntropy}
    & L(I^C_{v_i},I^C_{v_j},\hat{S}^C) = \\
    &\nonumber\;- \sum_{x \in C} \left\{ \alpha \hat{s}_x \log p_x + (1-\alpha)(1-\hat{s}_x)\log (1-p_x) \right\}.
\end{align}
For updating the weights of the model, we use stochastic gradient descent with Nesterov momentum update.
\par

Due to the relatively small number of 3D models in the dataset, we perform data augmentation in order to reduce overfitting and improve the generalization. Each cube $C$ is randomly translated and rotated and the color is varied by changing illumination and introducing Gaussian noise.


\subsection{Inference}\label{inference}
For inference, we process the scene not at once due to limitations of the GPU memory but divide the volume into cubes. For each cube, we first compute for both camera views the colored voxel cubes $I^C_{v_i}$ and $I^C_{v_j}$, which encode the camera parameters, and infer the surface probability $p_x$ for each voxel by the SurfaceNet. Since some of the cubes might not contain any surface voxels, we discuss in Section~\ref{implementation} how these empty cubes can be rejected in an early stage.

In order to convert the probabilities into a surface, we perform two operations. The first operation is a simple thresholding operation that converts all voxels with $p_x > \tau$ into surface voxels and all other voxels are set to zero. In Section \ref{adaptivethres}, we discuss how the threshold $\tau$ can be adaptively set when the 3D surface is recovered. The second operation is optional and it performs a thinning procedure of the surface since the surface might be several voxels thick after the binarization. To obtain a thin surface, we perform a pooling operation, which we call \textit{ray pooling}. For each view, we vote for a surface voxel $s_x=1$ if $x=\arg\underset{x' \in R}{\max}\; p_{x'}$, where $R$ denotes the voxels that are projected onto the same pixel. If both operations are used, a voxel $x$ is converted into a surface voxel if both views vote for it during ray pooling and $p_x > \tau$.

\section{Multi-View Stereopsis}\label{multiViewFusion}
So far we have described training and inference with SurfaceNet if only two views are available. We now describe how it can be trained and used for multi-view stereopsis.

\subsection{Inference}\label{multiViewFusion:inf}
If multiple views $v_1, \ldots, v_V$ are available, we select a subset of view pairs $(v_i,v_j)$ and compute for a cube $C$ and each selected view $v$ the CVC cube $I^C_{v}$. We will discuss at the end of the section how the view pairs are selected.

For each view pair $(v_i,v_j)$, SurfaceNet predicts $p^{(v_i,v_j)}_x$, \ie the confidence that a voxel $x$ is on the surface. The predictions of all view pairs can be combined by taking the average of the predictions $p^{(v_i,v_j)}_x$ for each voxel. However, the view pairs should not be treated equally since the reconstruction accuracy varies among the view pairs. In general, the accuracy depends on the viewpoint difference of the two views and the presence of occlusions. 

To further identify occlusions between two views $v_i$ and $v_j$, we crop a $64\times64$ patch around the projected center voxel of $C$ for each image $I_{v_i}$ and $I_{v_j}$. 
To compare the similarity of the two patches, we train a triplet network~\cite{schroff2015facenet} that learns a mapping $e(\cdot)$ from images to a compact 128D Euclidean space 
    where distances directly correspond to a measure of image similarity. 
The dissimilarity of the two patches is then given by
\begin{equation} \label{eq:similarity}
    d_C^{(v_i,v_j)} = \lVert e(C,I_{v_i})-e(C,I_{v_j}) \rVert_2,
\end{equation}
where $e(C,I_{v_i})$ denotes the feature embedding provided by the triplet network for the patch from image $I_{v_i}$. 
This measurement can be combined by the relation of the two viewpoints $v_i$ and $v_j$, which is measured by the angle between the projection rays of the center voxel of $C$, which is denoted by $\theta_C^{(v_i,v_j)}$.
We use a 2-layer fully connected neural network $r(\cdot)$, that has 100 hidden neurons with sigmoid activation function and one linear output neuron followed by a softmax layer. The relative weights for each view pair are then given by
\begin{equation} \label{eq:relativeImportance}
 w_C^{(v_i,v_j)} = r\left(\theta_C^{(v_i,v_j)}, d_C^{(v_i,v_j)}, e(C,I_{v_i})^T, e(C,I_{v_j})^T \right)
\end{equation}
and the weighted average of the predicted surface probabilities $p_x^{(v_i, v_j)}$ by 		
\begin{equation} \label{eq:fusion}
    p_x = \frac{\sum_{(v_i,v_j) \in \bm{V}_C}w_C^{(v_i,v_j)} p_x^{(v_i, v_j)}}
        {\sum_{(v_i,v_j) \in \bm{V}_C}w_C^{(v_i,v_j)}}
\end{equation}
where $\bm{V}_c$ denotes the set of selected view pairs. Since it is unnecessary to take all view pairs into account, we select only $N_v$ view pairs, which have the highest weight $w_C^{(v_i,v_j)}$. In Section \ref{experiments}, we evaluate the impact of $N_v$.

The binarization and thinning are performed as in Section \ref{inference}, \ie  a voxel $x$ is converted into a surface voxel if at least $\gamma=80\%$ of all views vote for it during ray pooling and $p_x > \tau$.
The effects of $\gamma$ and $\tau$ are further elaborated in Section \ref{experiments}.

\subsection{Training}
The SurfaceNet can be trained together with the averaging of multiple view pairs~(\ref{eq:fusion}). We select for each cube $C$, $N_v^{train}$ random view pairs and the loss is computed after the averaging of all view pairs. We use $N_v^{train}=6$ as a trade-off since larger values increase the memory for each sampled cube $C$ and thus require to reduce the batch size for training due to limited GPU memory. Note that for inference, $N_v$ can be larger or smaller than $N_v^{train}$.

In order to train the triplet network for the dissimilarity measurement $d_C^{(v_i,v_j)}$~(\ref{eq:similarity}), we sample cubes $C$ and three random views where the surface is not occluded. The corresponding patches obtained by projecting the center of the cube onto the first two views serve as a positive pair. The negative patch is obtained by randomly shifting the patch of the third view by at least a quarter of the patch size. While the first two views are different, the third view can be the same as one of the other two views.
At the same time, we use data augmentation by varying illumination or adding noise, rotation, scale, and translation.
After SurfaceNet and the triplet network are trained, we finally learn the shallow network $r(\cdot)$ (\ref{eq:relativeImportance}). 

\section{Implementation Details}\label{implementation}\label{adaptivethres}
As described in Section~\ref{inference}, the scene volume needs to be divided into cubes due to limitations of the GPU memory. However, it is only necessary to process cubes that are very likely to contain surface voxels. As an approximate measure, we apply logistic regression to the distance vector $d_C^{(v_i,v_j)}$~(\ref{eq:similarity}) for each view pair, which predicts if the patches of both views are similar. If for less than $N_{min}$ of the view pairs the predicted similarity probability is greater than or equal to $0.5$, we reject the cube. 


Instead of using a single threshold $\tau$ for binarization, one can also adapt the threshold for each cube $C$ based on its neighboring cubes $\mathcal{N}(C)$. The approach is iterated and we initialize $\tau_C$ by $0.5$ for each cube $C$. We optimize \mbox{$\tau_C \in [0.5,1)$} by minimizing the energy 
\begin{equation} \label{eq:E}
    E(\tau_C) = \sum_{C' \in \mathcal{N}(C)}{\psi\left(S^C(\tau_C), S^{C'}(\tau_{C'})\right)}, 
\end{equation}     
where $S^C(\tau_C)$ denotes the estimated surface in cube $C$ after binarization with threshold $\tau_C$.  

For the binary term $\psi$, we use 
\begin{equation}\label{eq:psi}
    \psi(S^C, S^{C'}) = \sum_{x \in C\cap C'} (1-s_x) s'_x + s_x (1-s'_x) - \beta s_x s'_x .
\end{equation}
The first two terms penalize if $S^C$ and $S^{C'}$ disagree in the overlapping region, which can be easily achieved by setting the threshold $\tau$ very high such that the overlapping region contains only very few surface voxels. The last term therefore aims to maximize the surface voxels that are shared among the cubes.       
\par
We used the \textit{Lasagne} Library \cite{lasagne} to implement the network structure. The code and the trained model are publically available.\footnote{\url{https://github.com/mjiUST/SurfaceNet}}
\section{Experiments} \label{experiments}

\subsection{Dataset}\label{dataset}

The DTU multi-view stereo dataset \cite{aanaes2016large} is a large scale MVS benchmark.
It features a variety of objects and materials, 
    and contains 80 different scenes seen from 49 or 64 camera positions 
    under seven different lighting conditions. The provided reference models are acquired by accurate structured light scans.
    The large selection of shapes and materials is well-suited to train and test our method under realistic conditions and the scenes are complex enough even for the state-of-the-art methods, 
    such as \textit{furu}~\cite{furu2010accurate}, \textit{camp}~\cite{campbell2008using}, \textit{tola}~\cite{tola2012efficient} and  \textit{Gipuma}~\cite{galliani2015massively}.  
For evaluation, we use three subsets of the objects from the DTU dataset
    for training, validation and evaluation.
\footnote{Training: 
        2, 6, 7, 8, 14, 16, 18, 19, 20, 22, 30, 31, 36, 39, 41, 42, 44, 45, 46, 47, 50, 51, 52, 53, 
        55, 57, 58, 60, 61, 63, 64, 65, 68, 69, 70, 71, 72, 74, 76, 83, 84, 85, 87, 88, 89, 90, 91, 
        92, 93, 94, 95, 96, 97, 98, 99, 100, 101, 102, 103, 104, 105, 107, 108, 109, 111, 112, 
        113, 115, 116, 119, 120, 121, 122, 123, 124, 125, 126, 127, 128. Validation: 3, 5, 17, 21, 28, 35, 37, 38, 40, 43, 56, 59, 66, 67, 82, 86, 106, 117. Evaluation: 1, 4, 9, 10, 11, 12, 13, 15, 23, 24, 29, 32, 33, 34, 48, 49, 62, 75, 77, 110, 114, 118}

The evaluation is based on \textit{accuracy} and \textit{completeness}. 
Accuracy is measured as the distance from the inferred model to the reference model, 
    and the completeness is calculated the other way around. Although both are actually error measures and a lower value is better, we use the terms \textit{accuracy} and \textit{completeness} as in \cite{aanaes2016large}.
Since the reference models of the DTU dataset are down-sampled to a resolution of $0.2mm$ for evaluation, 
we set the voxel resolution to $0.4mm$. 
In order to train SurfaceNet with a reasonable batch size, we use cubes with $32^3$ voxels.

\begin{figure}[htbp]
    \centering
    \newcommand{\colw}{0.30}
    \newcommand{\figw}{1.1} 
    \begin{subfigure}[b]{0.99\linewidth}
        \includegraphics[width=1.0\textwidth,trim={0.0cm 0.0cm 0.0cm 0.0cm},clip]{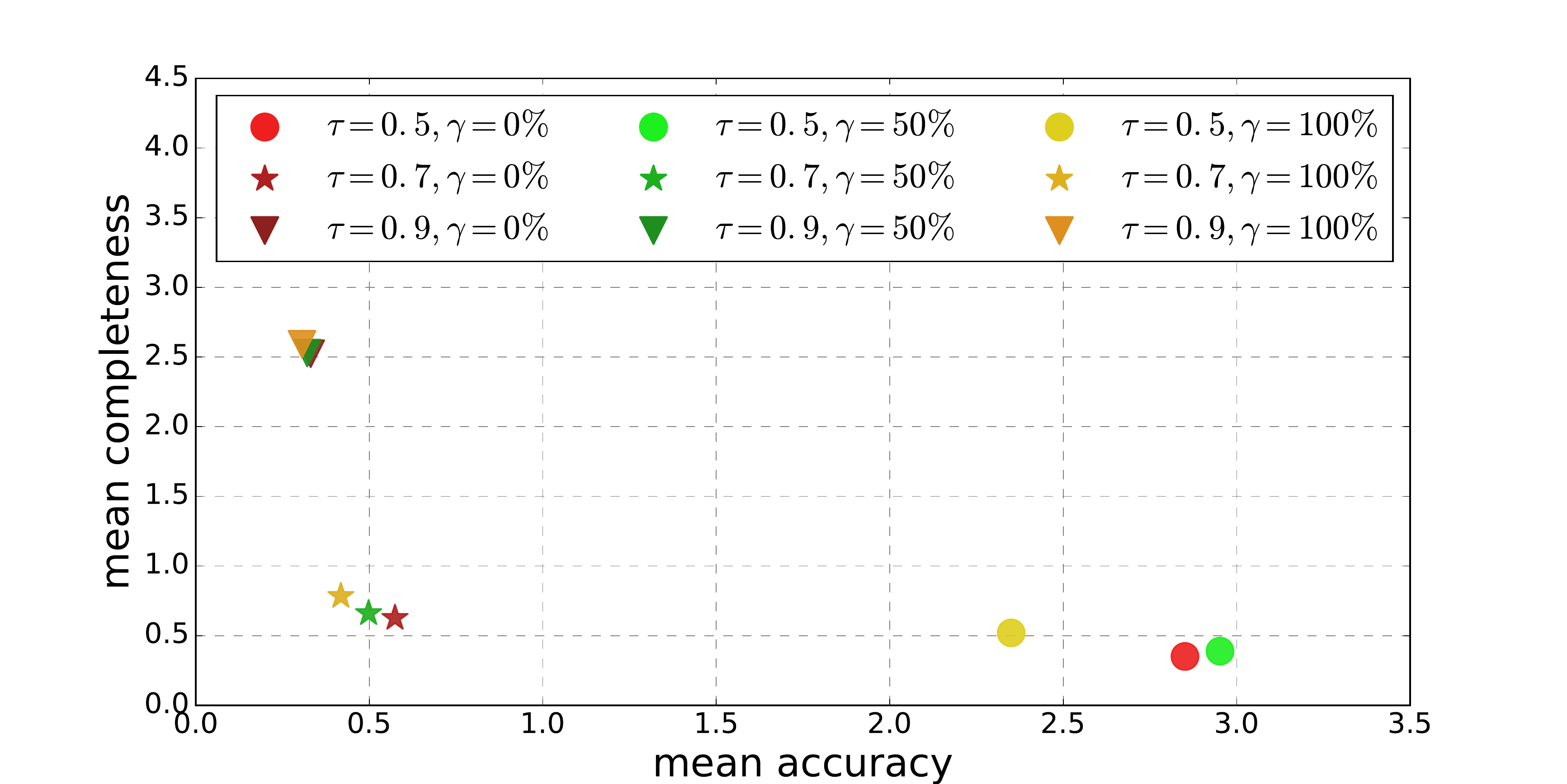}
    \end{subfigure}%

    \begin{subfigure}[t]{0.03\linewidth}
        \begin{minipage}{\textwidth}
            \vspace*{0.1cm}
            \rotatebox{90}{$\tau=0.5$}
        \end{minipage}
    \end{subfigure}%
    ~
    \begin{subfigure}[t]{\colw\linewidth}
        \begin{minipage}{\textwidth}
            \vspace*{0.1cm}
            \includegraphics[width=1\linewidth, keepaspectratio=true,trim={0cm 2cm 0cm 0cm},clip]{{{figures/results/model9/tau_gamma/t0.5_rp0}}}   
            \includegraphics[width=1\linewidth, keepaspectratio=true,trim={0cm 6cm 0cm 5cm},clip]{{{figures/results/model9/tau_gamma/sec-t0.5_rp0}}}
        \end{minipage}%
    \end{subfigure}%
    ~ 
    \begin{subfigure}[t]{\colw\linewidth}
        \begin{minipage}{\textwidth}
            \vspace*{0.1cm}
            \includegraphics[width=1\linewidth, keepaspectratio=true,trim={0cm 2cm 0cm 0cm},clip]{{{figures/results/model9/tau_gamma/t0.5_rp5}}}
            \includegraphics[width=1\linewidth, keepaspectratio=true,trim={0cm 6cm 0cm 5cm},clip]{{{figures/results/model9/tau_gamma/sec-t0.5_rp5}}}
        \end{minipage}
    \end{subfigure}
    ~ 
    \begin{subfigure}[t]{\colw\linewidth}
        \begin{minipage}{\textwidth}
            \vspace*{0.1cm}
            \includegraphics[width=1\linewidth, keepaspectratio=true,trim={0cm 2cm 0cm 0cm},clip]{{{figures/results/model9/tau_gamma/t0.5_rp10}}}
            \includegraphics[width=1\linewidth, keepaspectratio=true,trim={0cm 6cm 0cm 5cm},clip]{{{figures/results/model9/tau_gamma/sec-t0.5_rp10}}}
        \end{minipage}
    \end{subfigure}

    \begin{subfigure}[b]{0.03\linewidth}
        \begin{minipage}{\textwidth}
            \vspace*{0.1cm}
            \rotatebox{90}{$\tau=0.7$}
        \end{minipage}
    \end{subfigure}%
    ~
    \begin{subfigure}[t]{\colw\linewidth}
        \begin{minipage}{\textwidth}
            \vspace*{0.1cm}
            \includegraphics[width=1\linewidth,  keepaspectratio=true,trim={0cm 2cm 0cm 0cm},clip]{{{figures/results/model9/tau_gamma/t0.7_rp0}}}
            \includegraphics[width=1\linewidth,  keepaspectratio=true,trim={0cm 6cm 0cm 5cm},clip]{{{figures/results/model9/tau_gamma/sec-t0.7_rp0}}}
        \end{minipage}
    \end{subfigure}%
    ~ 
    \begin{subfigure}[t]{\colw\linewidth}
        \begin{minipage}{\textwidth}
            \vspace*{0.1cm}
            \includegraphics[width=1\linewidth,  keepaspectratio=true,trim={0cm 2cm 0cm 0cm},clip]{{{figures/results/model9/tau_gamma/t0.7_rp5}}}
            \includegraphics[width=1\linewidth,  keepaspectratio=true,trim={0cm 6cm 0cm 5cm},clip]{{{figures/results/model9/tau_gamma/sec-t0.7_rp5}}}
        \end{minipage}
    \end{subfigure}
    ~ 
    \begin{subfigure}[t]{\colw\linewidth}
        \begin{minipage}{\textwidth}
            \vspace*{0.1cm}
            \includegraphics[width=1\linewidth,  keepaspectratio=true,trim={0cm 2cm 0cm 0cm},clip]{{{figures/results/model9/tau_gamma/t0.7_rp10}}}
            \includegraphics[width=1\linewidth,  keepaspectratio=true,trim={0cm 6cm 0cm 5cm},clip]{{{figures/results/model9/tau_gamma/sec-t0.7_rp10}}}
        \end{minipage}
    \end{subfigure}

    \begin{subfigure}[b]{0.03\linewidth}
        \begin{minipage}{\textwidth}
            \vspace*{0.1cm}
            \rotatebox{90}{$\tau=0.9$}
        \end{minipage}
    \end{subfigure}%
    ~
    \begin{subfigure}[t]{\colw\linewidth}
        \begin{minipage}{\textwidth}
            \vspace*{0.1cm}
            \includegraphics[width=1\linewidth,  keepaspectratio=true,trim={0cm 2cm 0cm 0cm},clip]{{{figures/results/model9/tau_gamma/t0.9_rp0}}}
            \includegraphics[width=1\linewidth,  keepaspectratio=true,trim={0cm 6cm 0cm 5cm},clip]{{{figures/results/model9/tau_gamma/sec-t0.9_rp0}}}
        \end{minipage}
        \captionsetup{labelformat=empty}
        \caption{$\gamma=0\%$}
    \end{subfigure}%
    ~ 
    \begin{subfigure}[t]{\colw\linewidth}
        \begin{minipage}{\textwidth}
            \vspace*{0.1cm}
            \includegraphics[width=1\linewidth,  keepaspectratio=true,trim={0cm 2cm 0cm 0cm},clip]{{{figures/results/model9/tau_gamma/t0.9_rp5}}}
            \includegraphics[width=1\linewidth,  keepaspectratio=true,trim={0cm 6cm 0cm 5cm},clip]{{{figures/results/model9/tau_gamma/sec-t0.9_rp5}}}
        \end{minipage}
        \captionsetup{labelformat=empty}
        \caption{$\gamma=50\%$}
    \end{subfigure}
    ~ 
    \begin{subfigure}[t]{\colw\linewidth}
        \begin{minipage}{\textwidth}
            \vspace*{0.1cm}
            \includegraphics[width=1\linewidth,  keepaspectratio=true,trim={0cm 2cm 0cm 0cm},clip]{{{figures/results/model9/tau_gamma/t0.9_rp10}}}
            \includegraphics[width=1\linewidth,  keepaspectratio=true,trim={0cm 6cm 0cm 5cm},clip]{{{figures/results/model9/tau_gamma/sec-t0.9_rp10}}}
        \end{minipage}
        \captionsetup{labelformat=empty}
        \caption{$\gamma=100\%$}
    \end{subfigure}
    \caption{Quantitative and qualitative evaluation of $\tau$ and $\gamma$. 
        } 
   \label{fig:tau_gamma}
\end{figure}

\subsection{Impact of Parameters}

We first evaluate the impact of the parameters for binarization. In order to provide a quantitative and qualitative analysis, we use model 9 of the dataset, which was randomly chosen from the test set. By default, we use cubes with $32^3$ voxels.   
 
As discussed in Section \ref{multiViewFusion:inf}, the binarization depends on the threshold $\tau$ and the parameter $\gamma$ for thinning. If $\gamma=0\%$ thinning is not performed. Fig.~\ref{fig:tau_gamma} shows the impact of $\tau$ and $\gamma$. 
For each parameter setting, the front view and the intersection with a horizontal plane (red) is shown from top view. The top view shows the thickness and consistency of the reconstruction.
%
%
We observe that $\tau$ is a trade-off between accuracy and completeness. While a large value $\tau=0.9$ discards large parts of the surface, $\tau=0.5$ results in a noisy and inaccurate reconstruction. The thinning, i.e., $\gamma=100\%$, improves the accuracy for any threshold $\tau$ and slightly impacts the completeness.

\begin{figure}[htbp]
    \centering
    \newcommand{\colw}{0.45}
    \newcommand{\figw}{1.0} 
    \begin{subfigure}[b]{0.8\linewidth}
        \includegraphics[width=\figw\textwidth,trim={5cm 8cm 5.5cm 7cm},clip]{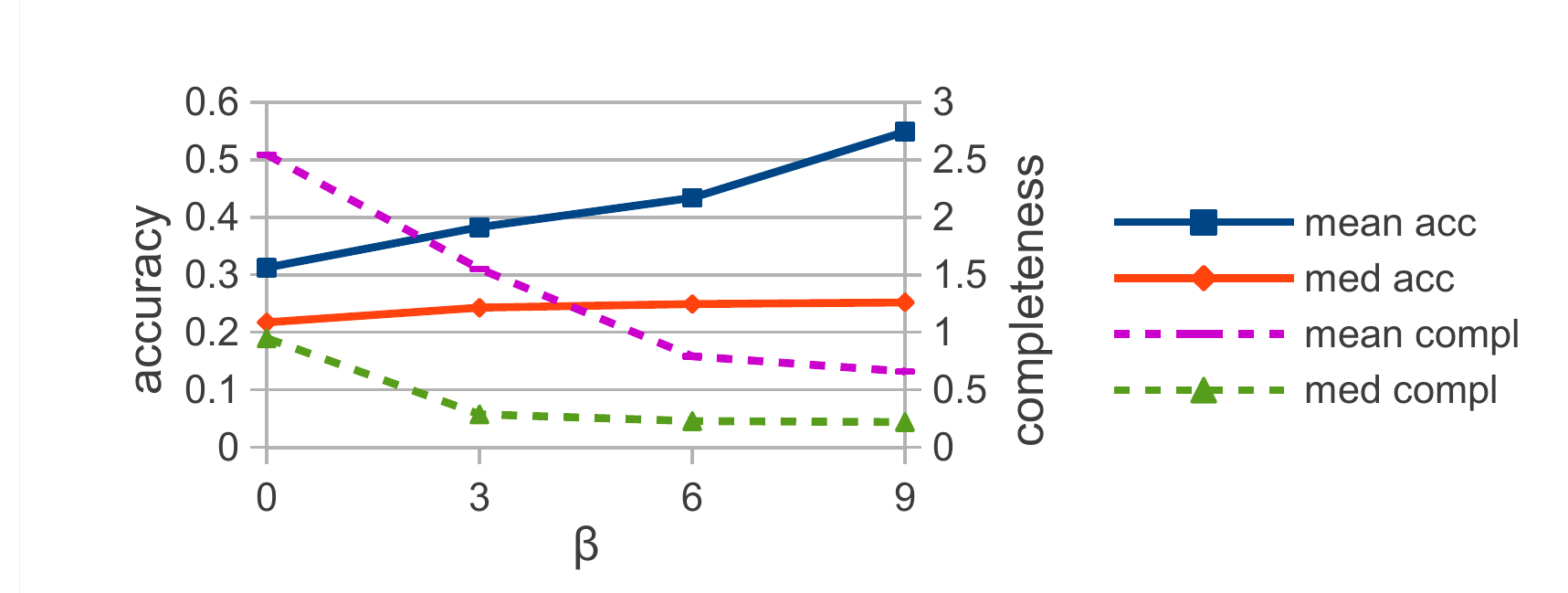}  
    \end{subfigure}%

    
    \begin{subfigure}[b]{\colw\linewidth}
        \includegraphics[width=\figw\textwidth,trim={0cm 3cm 0cm 3cm},clip]{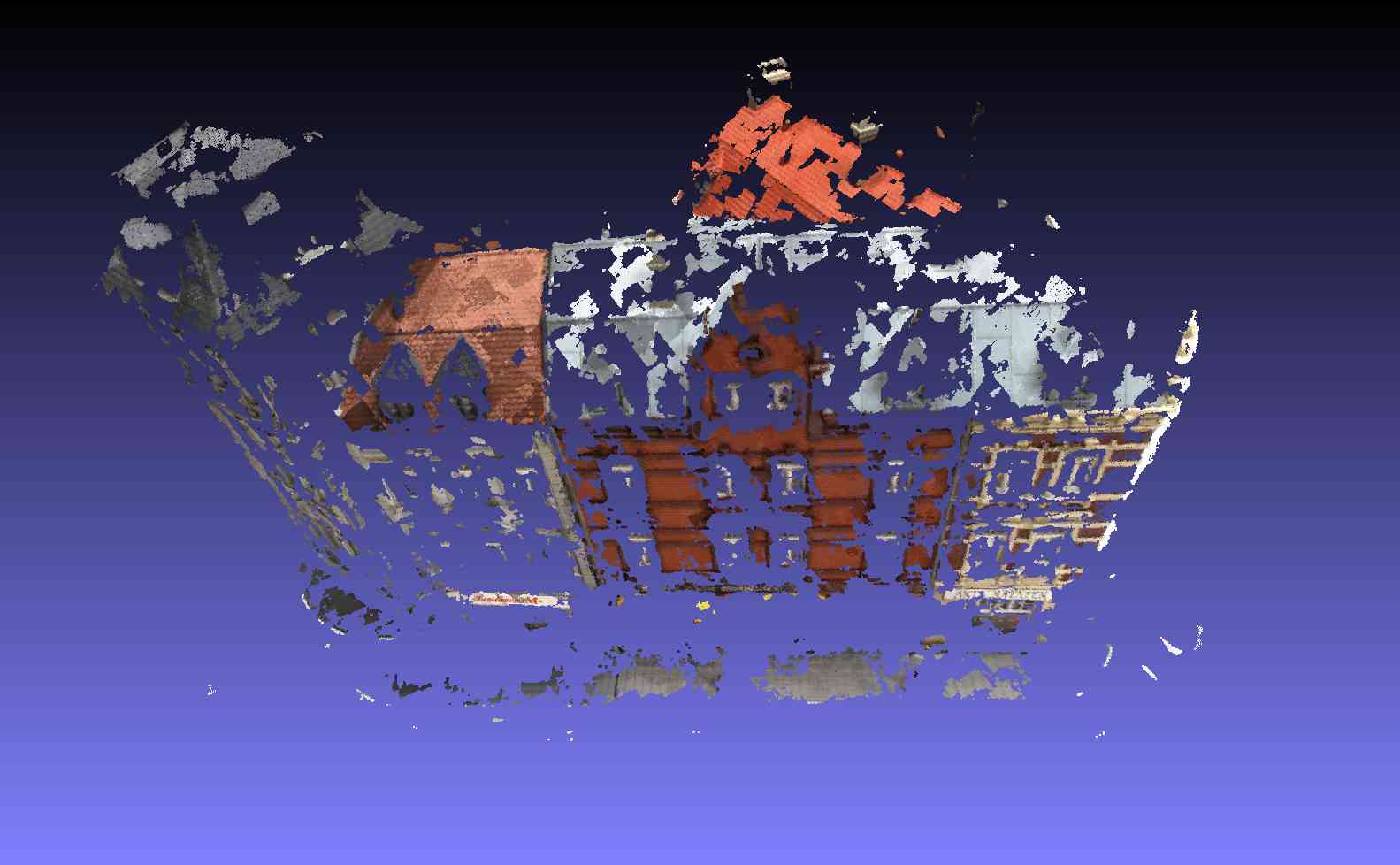} 
        \captionsetup{labelformat=empty}
        \caption{$\beta = 0$}
    \end{subfigure}
    ~ 
    \begin{subfigure}[b]{\colw\linewidth}
        \includegraphics[width=\figw\textwidth,trim={0cm 3cm 0cm 3cm},clip]{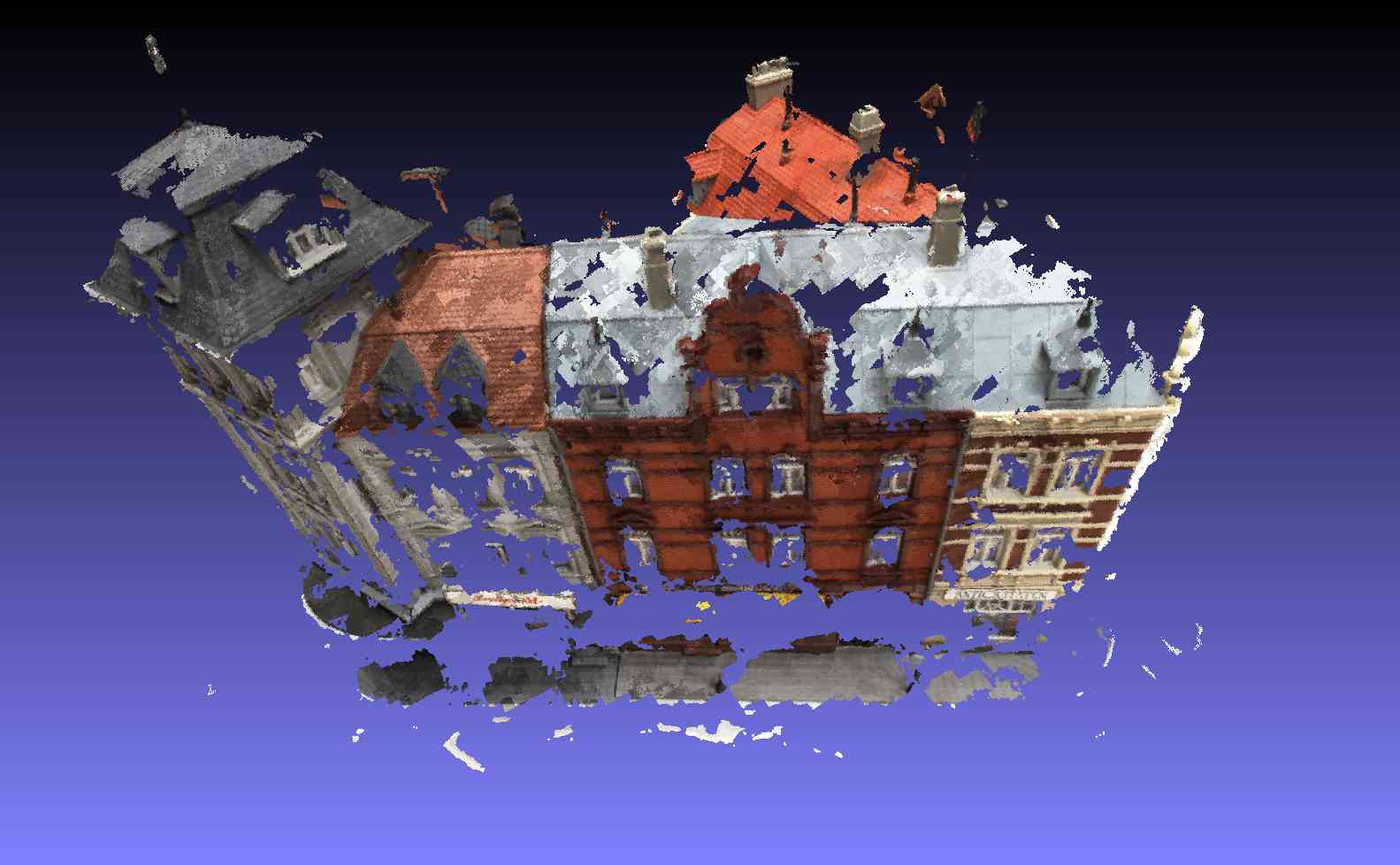}
        \captionsetup{labelformat=empty}
        \caption{$\beta = 3$}
    \end{subfigure}
    ~ 
    \begin{subfigure}[b]{\colw\linewidth}
        \includegraphics[width=\figw\textwidth,trim={0cm 3cm 0cm 3cm},clip]{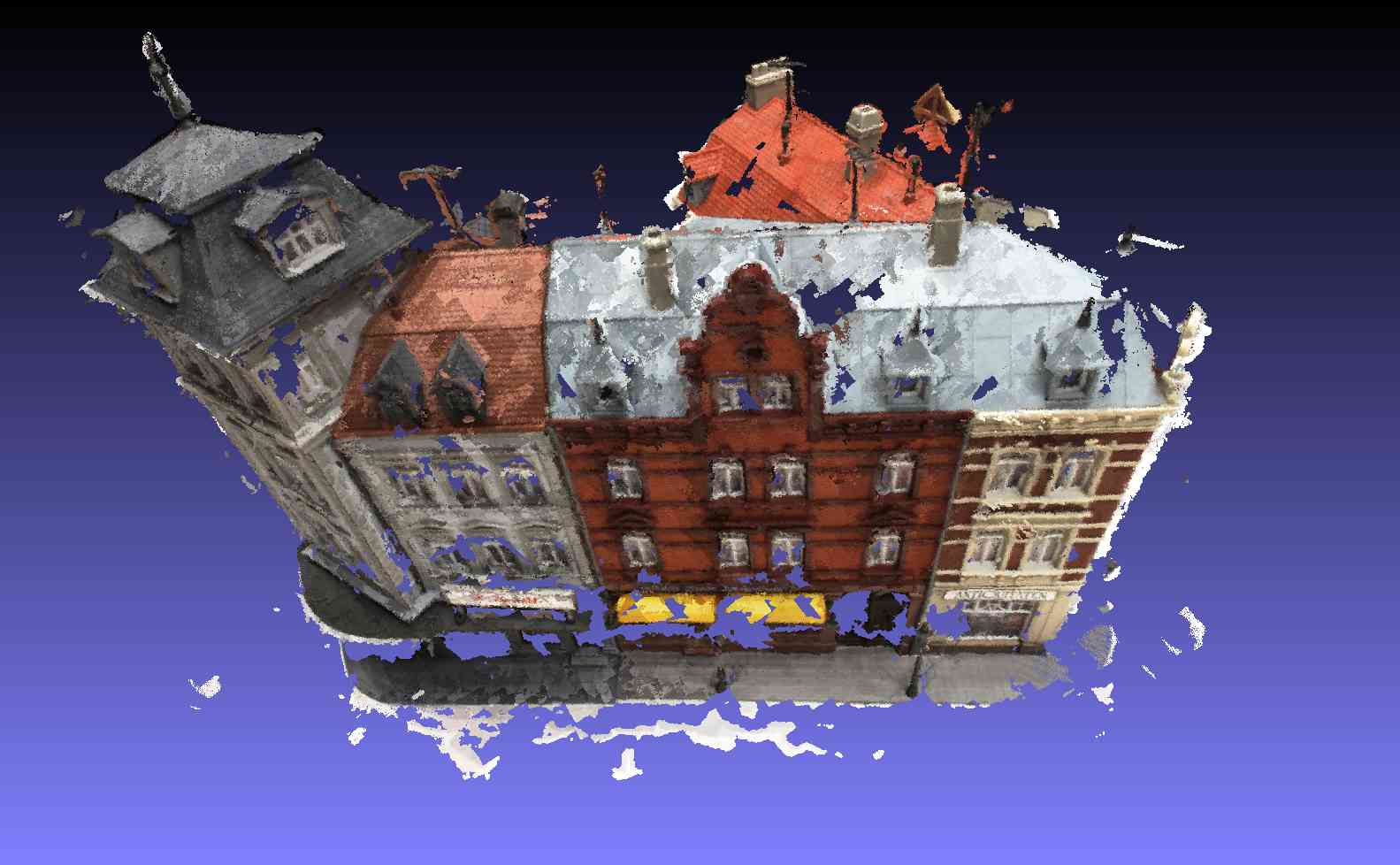}
        \captionsetup{labelformat=empty}
        \caption{$\beta = 6$}
    \end{subfigure}
    ~ 
    \begin{subfigure}[b]{\colw\linewidth}
        \includegraphics[width=\figw\textwidth,trim={0cm 3cm 0cm 3cm},clip]{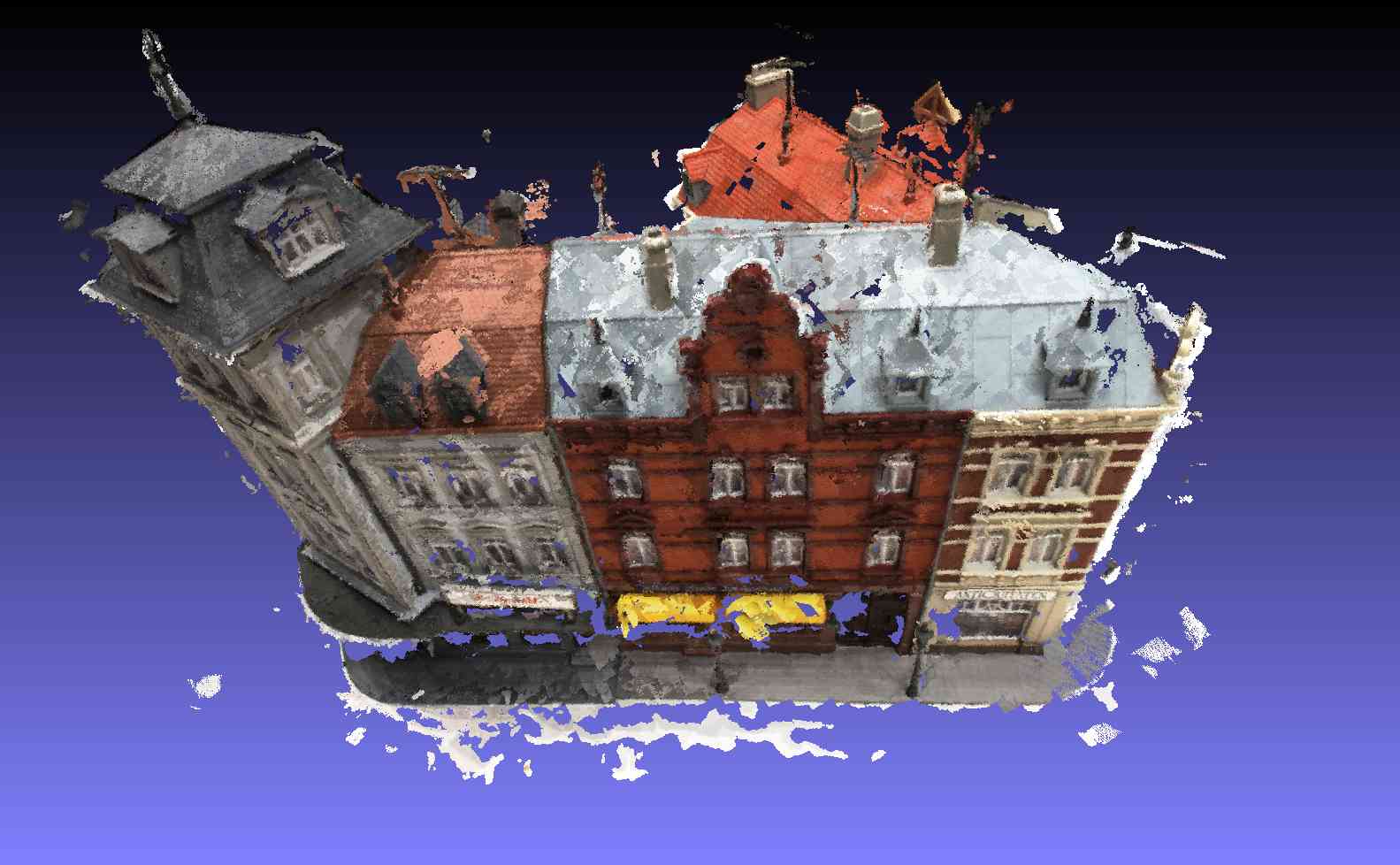}
        \captionsetup{labelformat=empty}
        \caption{$\beta = 9$}
    \end{subfigure}
    \caption{
    Quantitative and qualitative evaluation of $\beta$.
}
\label{fig:beta}
\end{figure}

\begin{figure}[htbp]
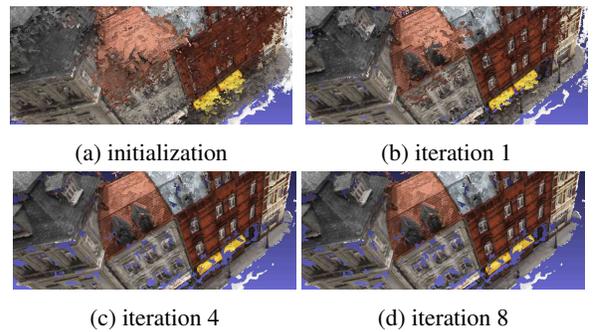

    \centering
    \newcommand{\colw}{0.45}
    \newcommand{\figw}{1.0} 
    \begin{subfigure}[t]{\colw\linewidth}
        \includegraphics[width=\figw\textwidth,trim={1cm 5cm 10cm 8cm},clip]{{{figures/results/model9/iter/iter0}}}
        \caption{initialization}
    \end{subfigure}%
    ~ 
    \begin{subfigure}[t]{\colw\linewidth}
        \includegraphics[width=\figw\textwidth,trim={1cm 5cm 10cm 8cm},clip]{{{figures/results/model9/iter/iter1}}}
        \caption{iteration 1}
    \end{subfigure}
    ~ 
    \begin{subfigure}[t]{\colw\linewidth}
        \includegraphics[width=\figw\textwidth,trim={1cm 5cm 10cm 8cm},clip]{{{figures/results/model9/iter/iter4}}}
        \caption{iteration 4}
    \end{subfigure}
    \begin{subfigure}[t]{\colw\linewidth}
        \includegraphics[width=\figw\textwidth,trim={1cm 5cm 10cm 8cm},clip]{{{figures/results/model9/iter/iter8}}}
        \caption{iteration 8}
    \end{subfigure}%
    \caption{
        The adaptive threshold is estimated by an iterative algorithm.
        The algorithm converges within a few iterations.
    }
   \label{fig:iter}
\end{figure}

While the threshold $\tau=0.7$ seems to provide a good trade-off between accuracy and completeness, we also evaluate the approach described in Section \ref{adaptivethres} where the constant threshold $\tau$ is replaced by an adaptive threshold $\tau_C$ that is estimated for each cube by minimizing~(\ref{eq:E}) iteratively. The energy, however, also provides a trade-off parameter $\beta$~(\ref{eq:psi}). If $\beta$ is large, we prefer completeness and when $\beta$ is small we prefer accuracy.      
This is reflected in Fig.~\ref{fig:beta} where we show the impact of $\beta$. 
Fig.~\ref{fig:iter} shows how the reconstruction improves for $\beta=6$ with the number of iterations from the initialization with $\tau_C=0.5$. The method quickly converges after a few iterations. By default, we use $\beta=6$ and 8 iterations for the adaptive binarization approach.

%
%

\begin{table}[htbp]
\centering
\small
\begin{tabular}{ccccc}
methods (mm) & \begin{tabular}{@{}c@{}}mean \\ acc\end{tabular}
        & \begin{tabular}{@{}c@{}}med \\ acc\end{tabular}
        & \begin{tabular}{@{}c@{}}mean \\ compl\end{tabular}
        & \begin{tabular}{@{}c@{}}med \\ compl\end{tabular} \\
    \hline
    $\tau=0.7$ $\gamma=0\%$ &             0.574	&0.284    &\textbf{0.627}	&\textbf{0.202} \\
    \hline
    $\tau=0.7$ $\gamma=80\%$ &          0.448	&\textbf{0.234}	&0.706	&0.242 \\
    \hline
    adaptive threshold \\ $\beta=6$ $\gamma=80\%$ &   \textbf{0.434}	&0.249	&0.792	&0.229 \\
    \hline
    adaptive threshold \\ $\beta=6$ $\gamma=80\%$ \\ w/o weighted average &  0.448 &   0.251 &       0.798 & 0.228 \\
    \hline
\end{tabular}
\captionof{table}{
A quantitative comparison of two well performing parameter settings for $\tau$ and $\gamma$ with the adaptive binarization procedure. The last row reports the result when the view pairs are not weighted.
    The evaluation is performed for the model~9.
} \label{tab:adapthresh}
\end{table}

We compare the adaptive binarization with the constant thresholding in Table~\ref{tab:adapthresh} where we report mean and median accuracy and completeness for model 9. The configuration $\tau=0.7$ and $\gamma=80\%$ provides a good trade-off between accuracy and completeness. Using adaptive thresholding with $\beta=6$ as described in Section~\ref{adaptivethres} achieves a better mean accuracy but the mean completeness is slightly worse. We also report in the last row the result when the probabilities $p_x$ are not weighted in (\ref{eq:fusion}), \ie, $w_C^{(v_i,v_j)}=1$. This slightly deteriorates the accuracy.


\begin{figure}[htbp]
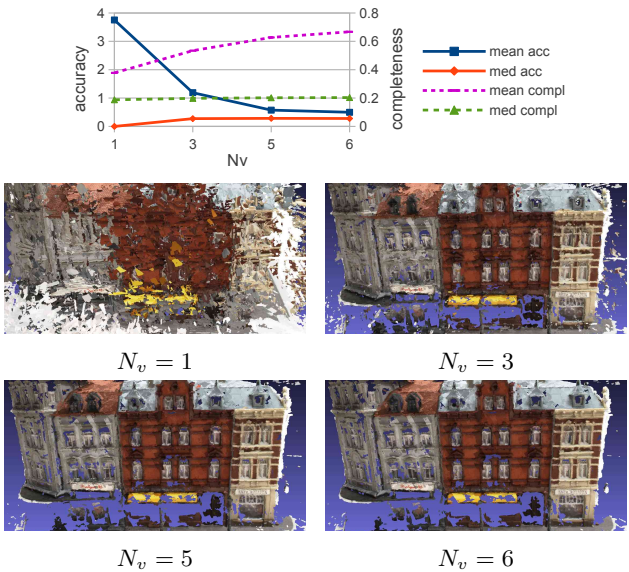

    \centering
    \newcommand{\colw}{0.48}
    \newcommand{\figw}{1.0} 
    \begin{subfigure}[b]{0.8\linewidth}
        \includegraphics[width=1.0\textwidth,trim={6cm 8cm 6cm 8cm},clip]{{{figures/table/Nv}}}
    \end{subfigure}%
    
    \begin{subfigure}[b]{\colw\linewidth}
        \includegraphics[width=\figw\textwidth,trim={5cm 5cm 7cm 6cm},clip]{{{figures/results/model9/Nv/1}}}
        \captionsetup{labelformat=empty}
        \caption{$N_v=1$}
    \end{subfigure}
    ~ 
    \begin{subfigure}[b]{\colw\linewidth}
        \includegraphics[width=\figw\textwidth,trim={5cm 5cm 7cm 6cm},clip]{{{figures/results/model9/Nv/3}}}
        \captionsetup{labelformat=empty}
        \caption{$N_v=3$}
    \end{subfigure}
    ~ 
    \begin{subfigure}[b]{\colw\linewidth}
        \includegraphics[width=\figw\textwidth,trim={5cm 5cm 7cm 6cm},clip]{{{figures/results/model9/Nv/5}}}
        \captionsetup{labelformat=empty}
        \caption{$N_v=5$}
    \end{subfigure}
    ~ 
    \begin{subfigure}[b]{\colw\linewidth}
        \includegraphics[width=\figw\textwidth,trim={5cm 5cm 7cm 6cm},clip]{{{figures/results/model9/Nv/6}}}
        \captionsetup{labelformat=empty}
        \caption{$N_v=6$}
    \end{subfigure}
    \caption{Quantitative and qualitative evaluation of $N_v$. 
        } 
   \label{fig:Nv}
\end{figure}

We finally evaluate the impact of fusing the probabilities $p_x$ of the best $N_v$ view pairs in (\ref{eq:fusion}). By default, we used $N_v=5$ so far. The impact of $N_v$ is shown in Fig.~\ref{fig:Nv}. While taking only the best view pair results in a very noisy inaccurate reconstruction, the accuracy is substantially improved for three view pairs. After five view pairs the accuracy slightly improves at the cost of a slight deterioration of the completeness. We therefore keep $N_v=5$.          


\begin{figure*}[htbp]
    \centering
    \newcommand{\colw}{0.16}
    \newcommand{\figw}{1.0} 
    \addtocounter{subfigure}{-4}
    \begin{minipage}[b]{\colw\textwidth}
        \begin{overpic}[width=\figw\textwidth,trim={0cm 0cm 0cm 0cm},clip]{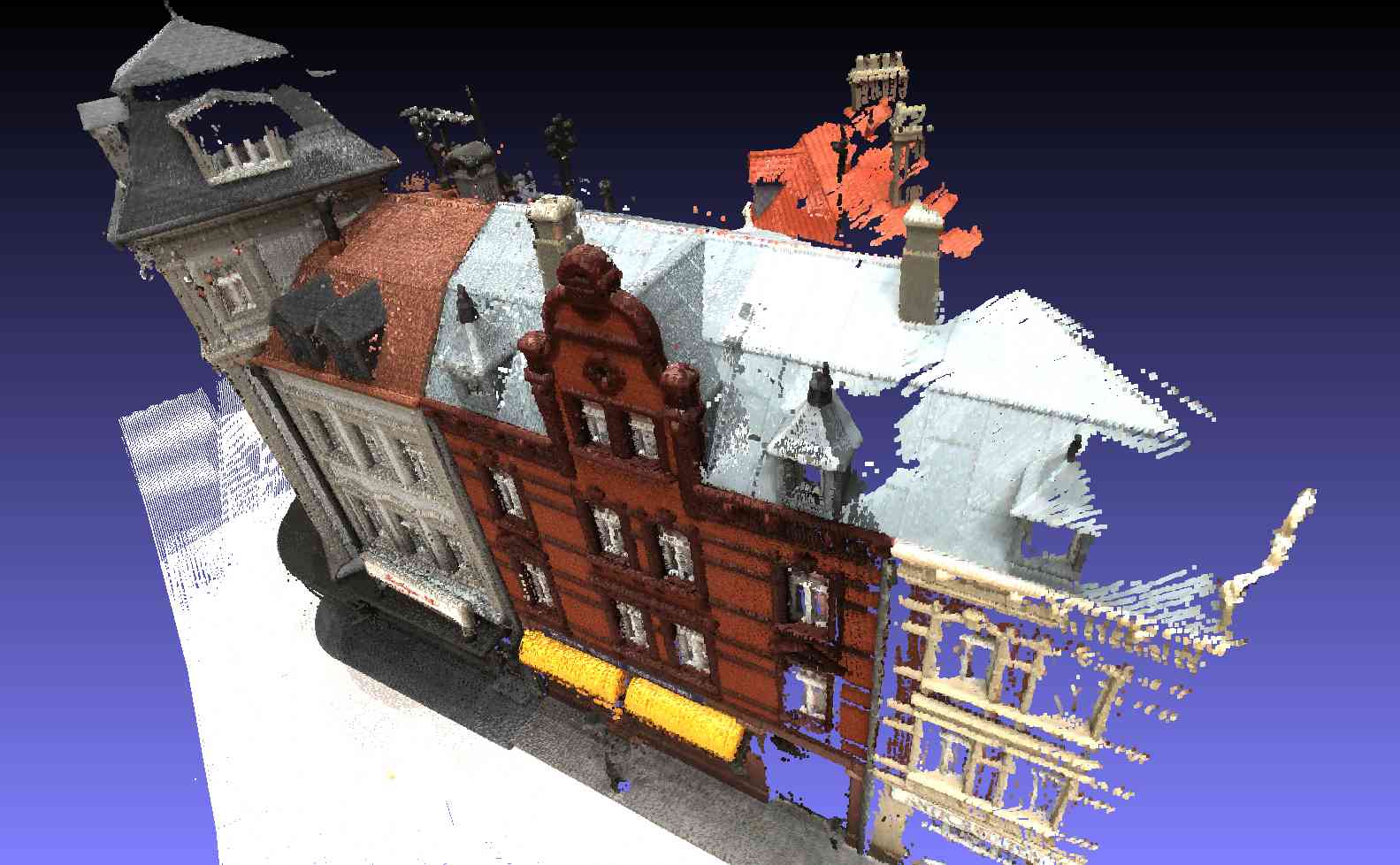}\end{overpic}
        \begin{overpic}[width=\figw\textwidth,trim={0cm 0cm 0cm 0cm},clip]{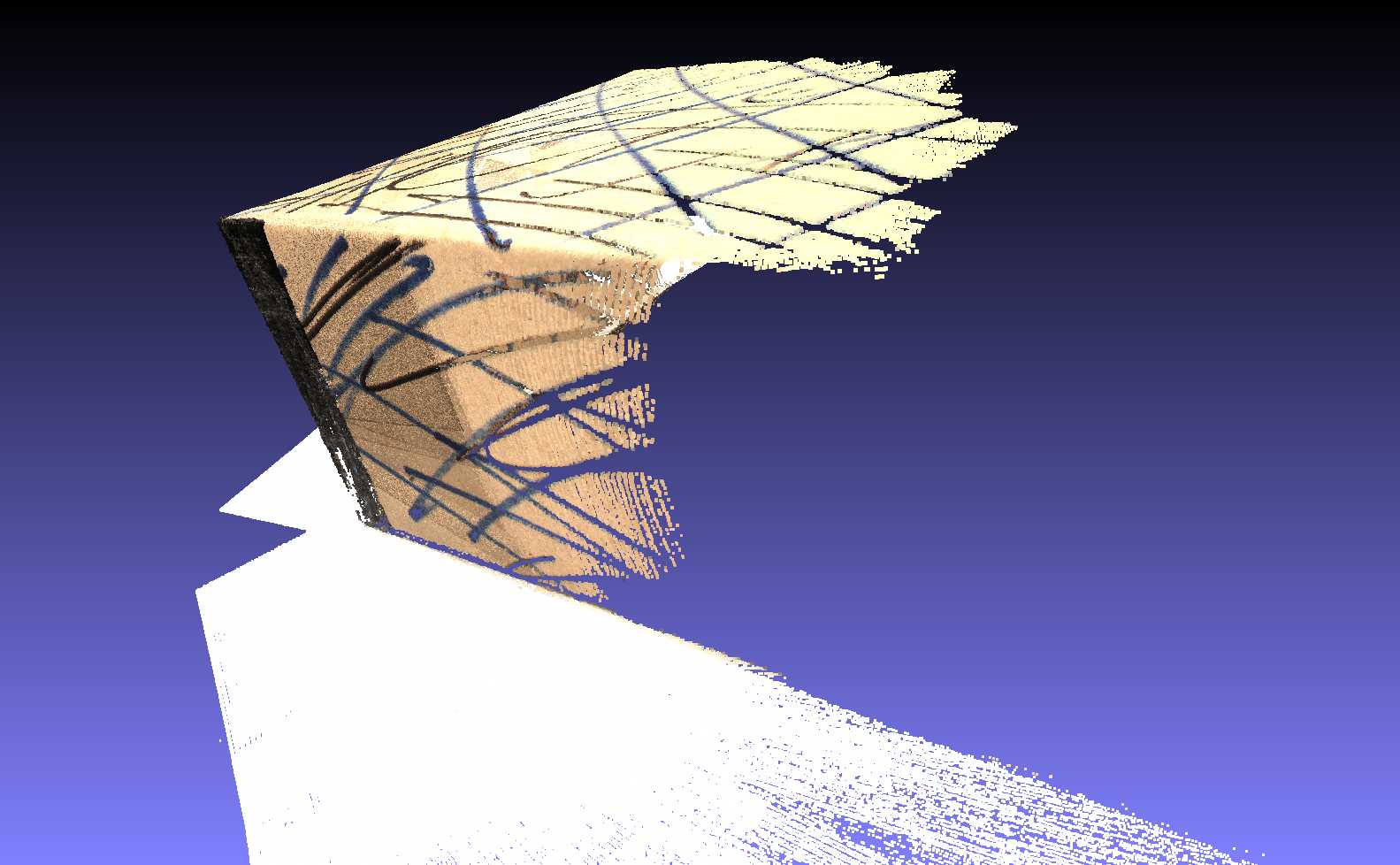}\end{overpic}
        \begin{overpic}[width=\figw\textwidth,trim={0cm 0cm 0cm 0cm},clip]{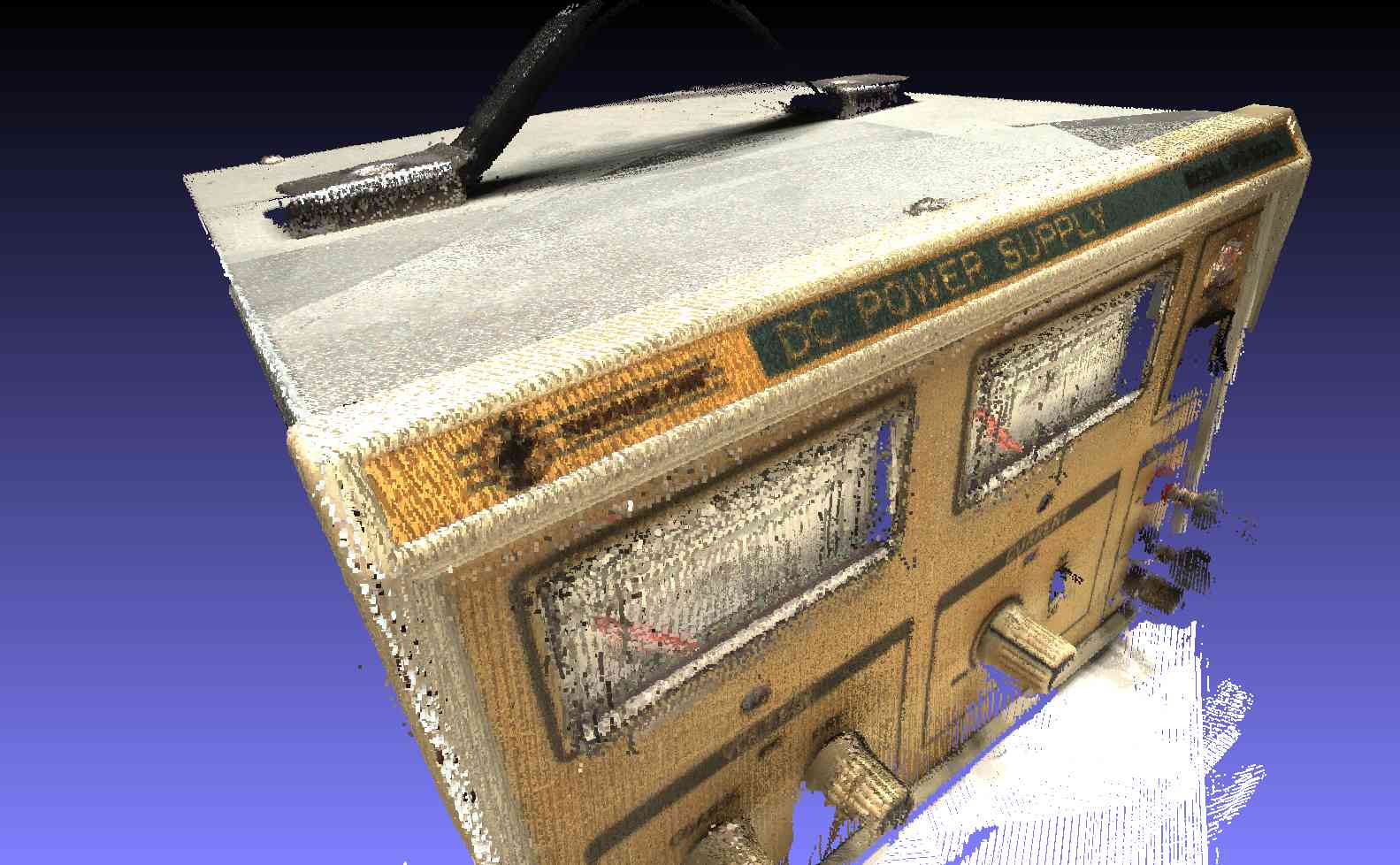}\end{overpic}
        \captionof{subfigure}{reference model}
    \end{minipage}
    \begin{minipage}[b]{\colw\textwidth}
        \begin{overpic}[width=\figw\textwidth,trim={0cm 0cm 0cm 0cm},clip]{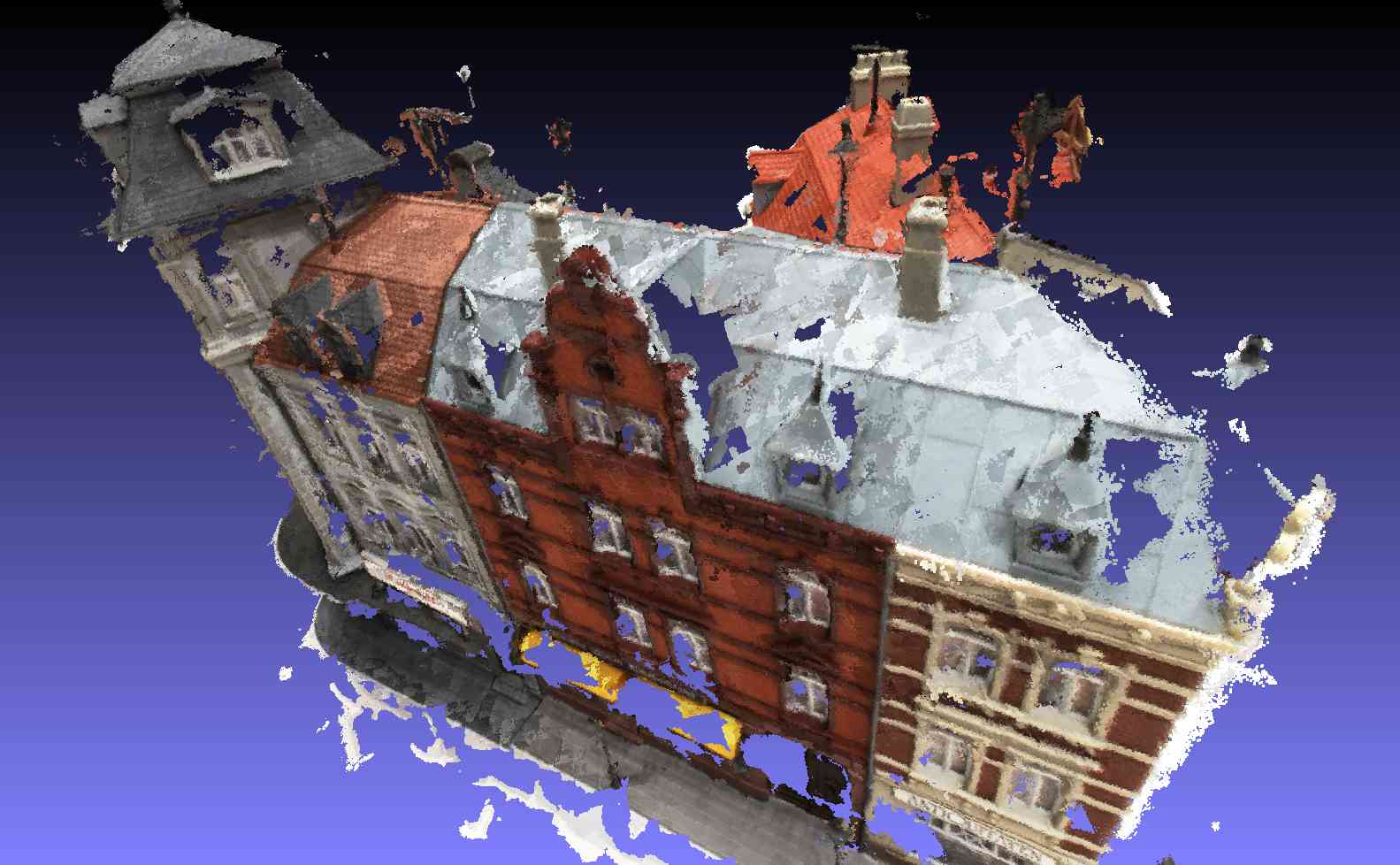}\end{overpic}
        \begin{overpic}[width=\figw\textwidth,trim={0cm 0cm 0cm 0cm},clip]{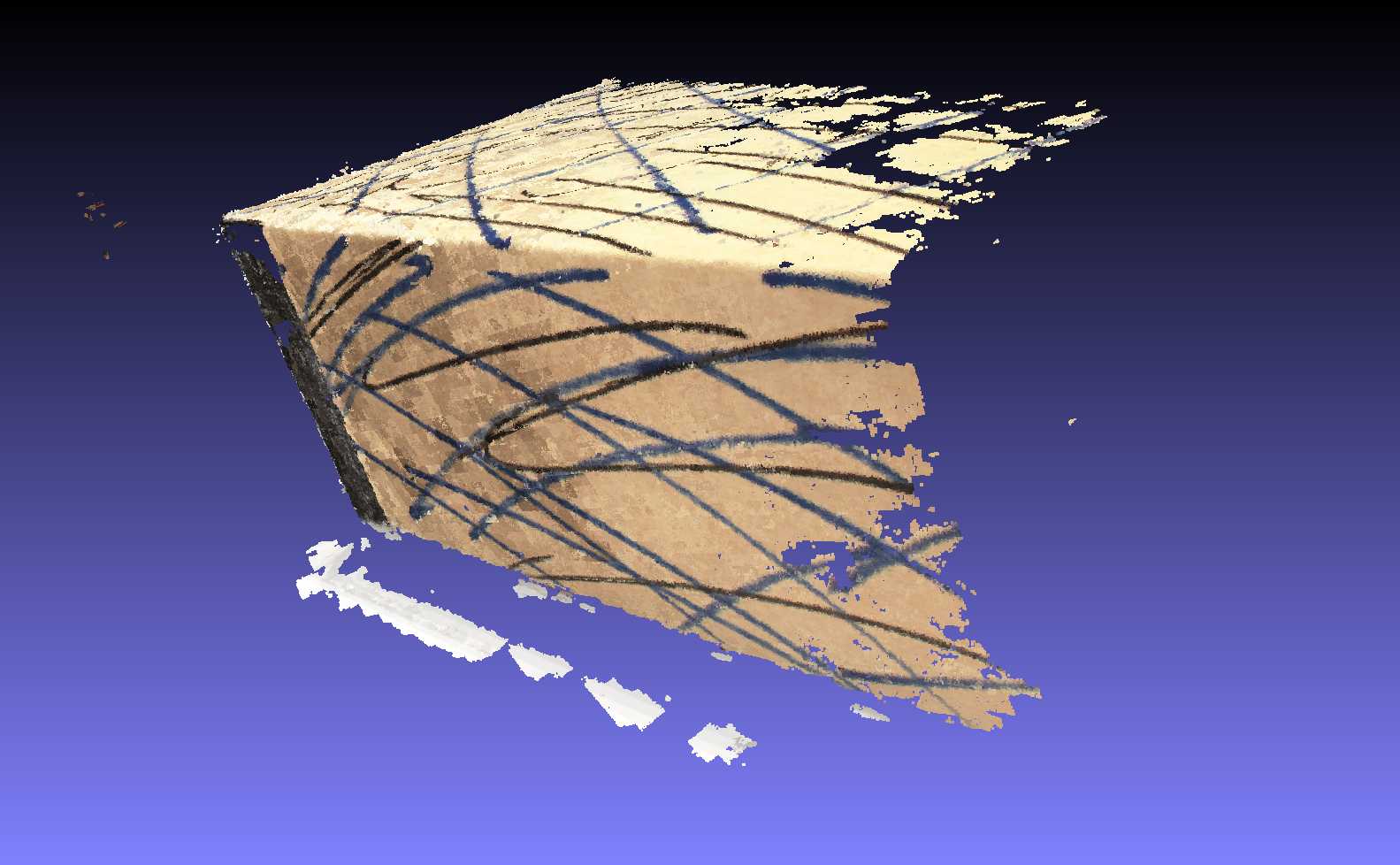}\end{overpic}
        \begin{overpic}[width=\figw\textwidth,trim={0cm 0cm 0cm 0cm},clip]{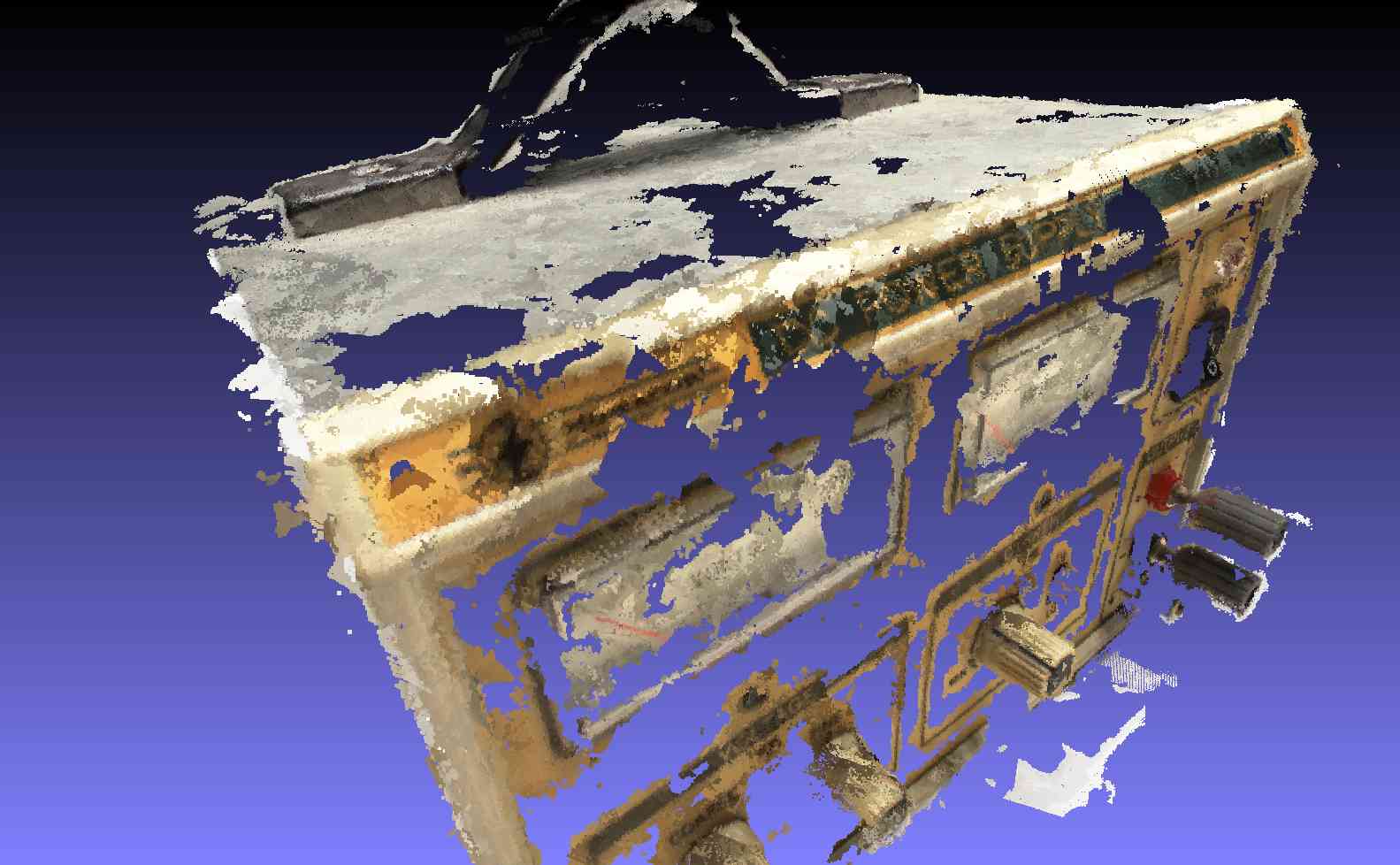}\end{overpic}
        \captionof{subfigure}{\textbf{SurfaceNet}}
    \end{minipage}
    \begin{minipage}[b]{\colw\textwidth}
        \begin{overpic}[width=\figw\textwidth,trim={0cm 0cm 0cm 0cm},clip]{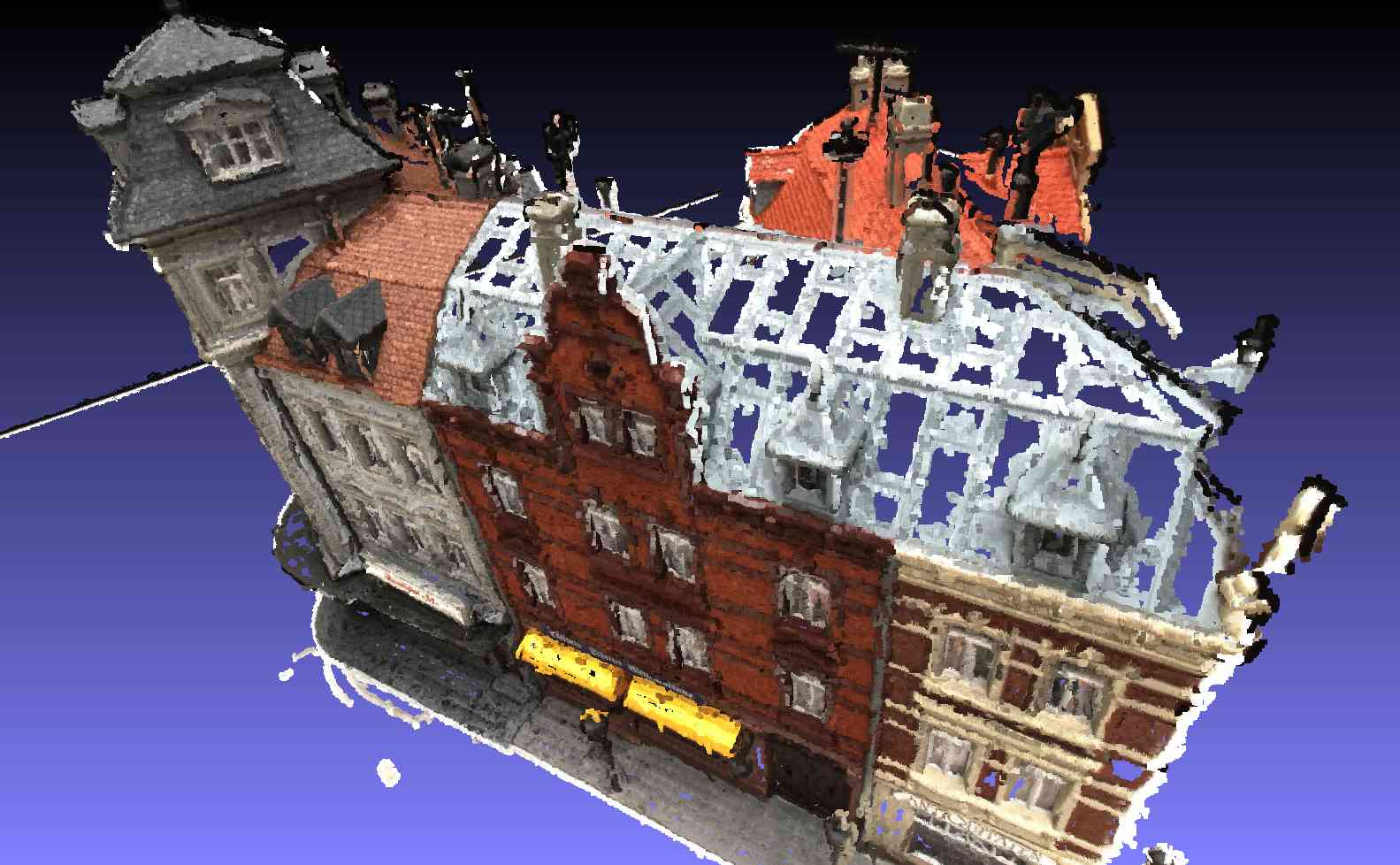}\end{overpic}
        \begin{overpic}[width=\figw\textwidth,trim={0cm 0cm 0cm 0cm},clip]{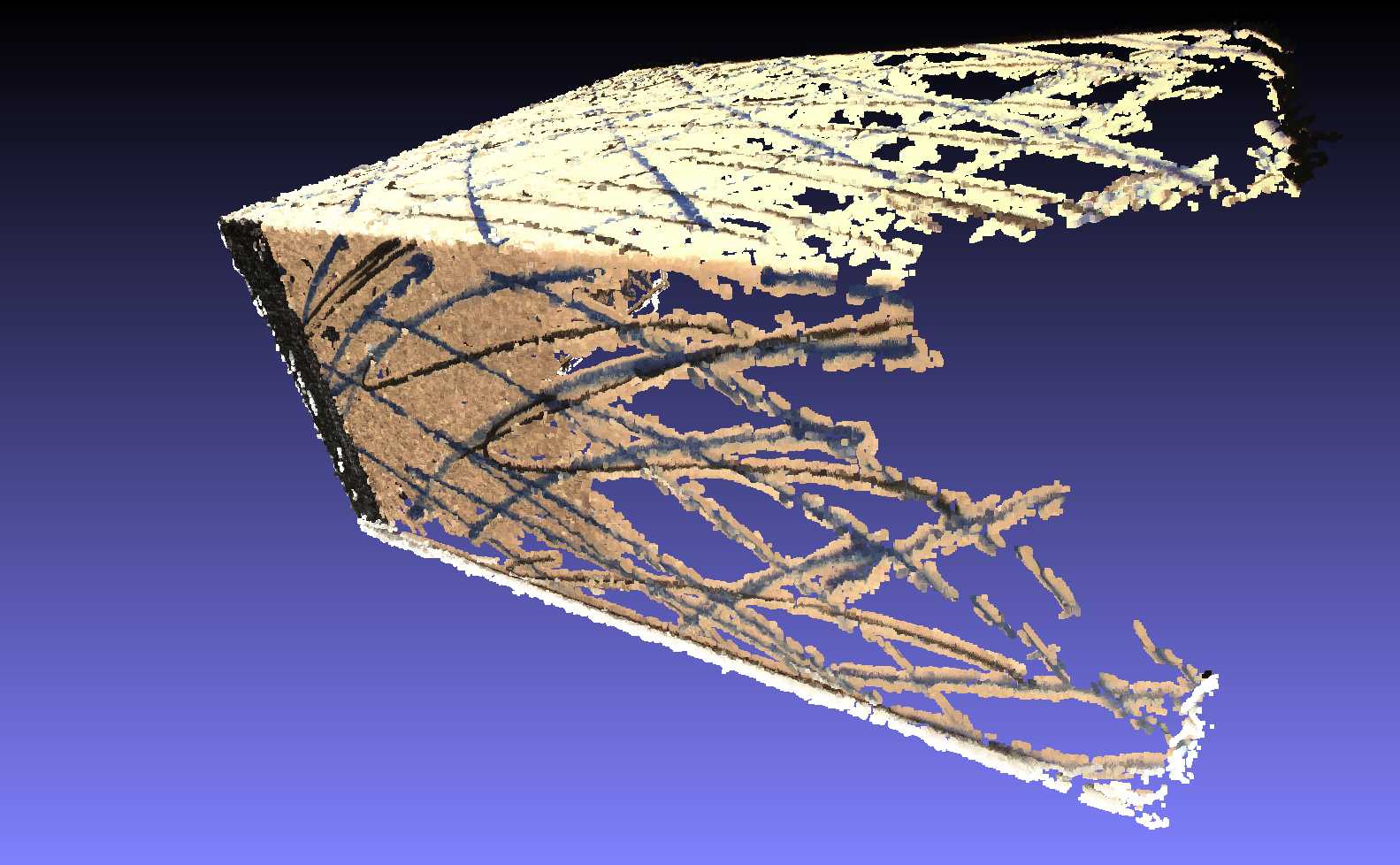}\end{overpic}
        \begin{overpic}[width=\figw\textwidth,trim={0cm 0cm 0cm 0cm},clip]{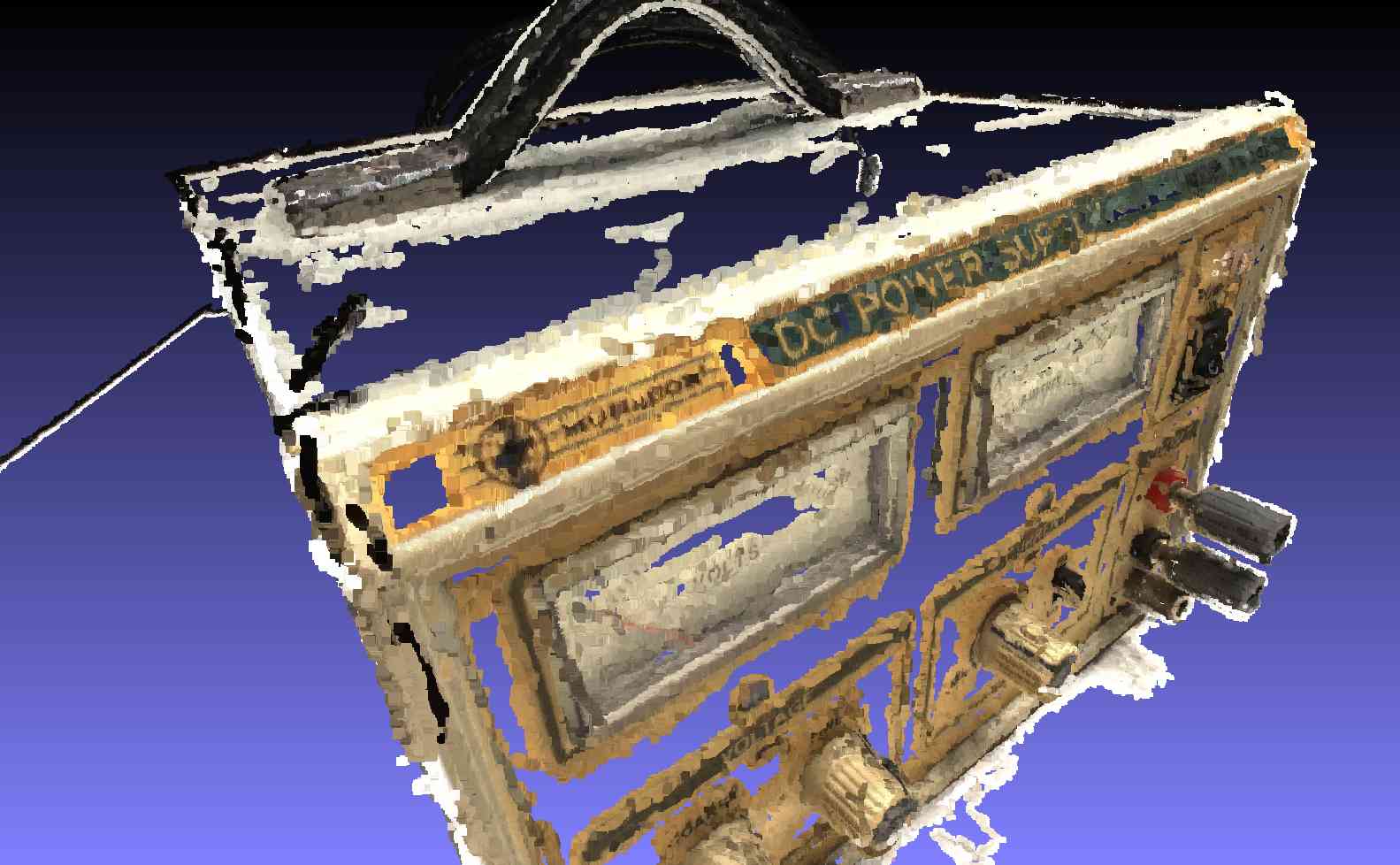}\end{overpic}
        \captionof{subfigure}{\textit{camp} \cite{campbell2008using}}
    \end{minipage}
    \begin{minipage}[b]{\colw\textwidth}
        \begin{overpic}[width=\figw\textwidth,trim={0cm 0cm 0cm 0cm},clip]{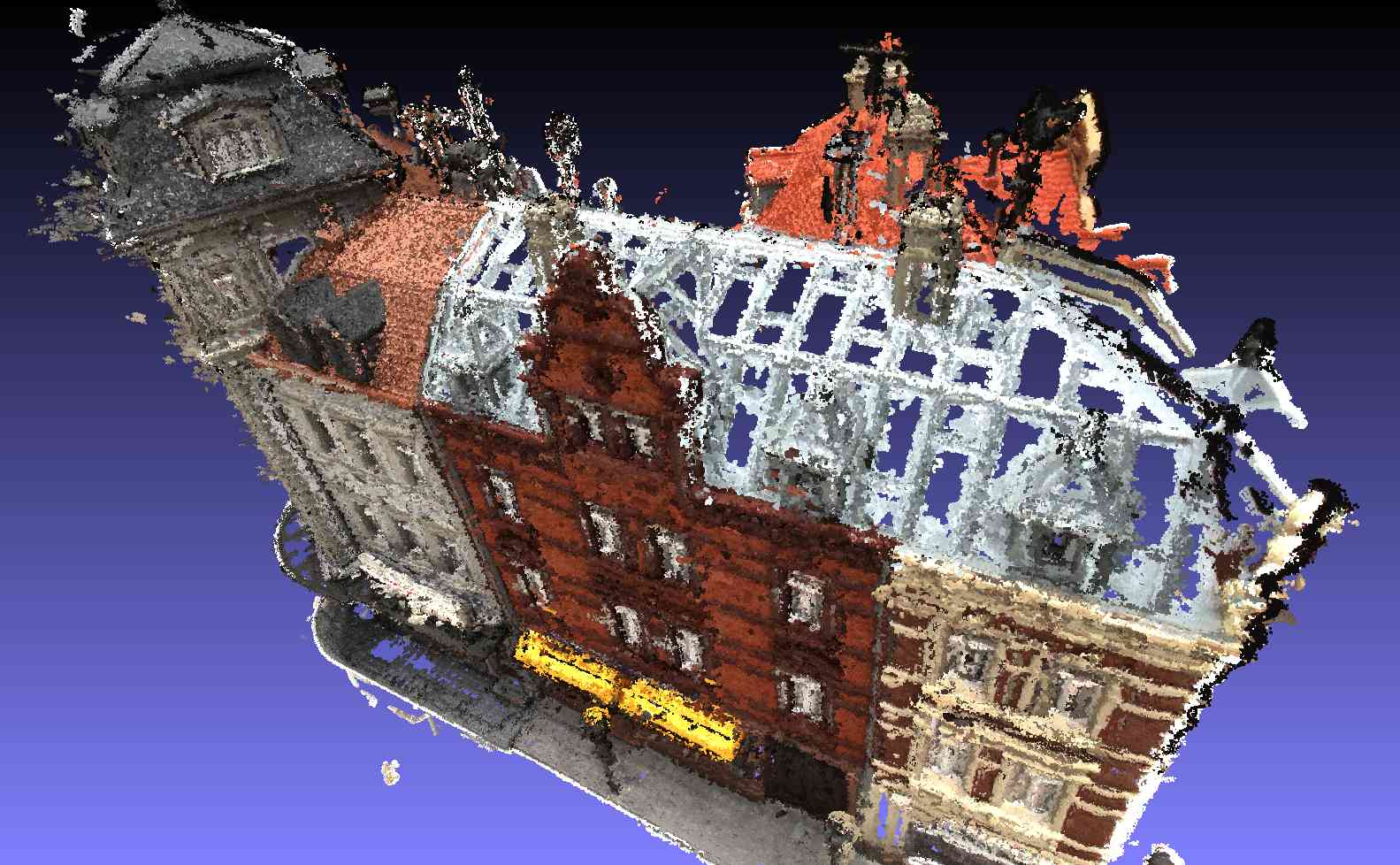}\end{overpic}
        \begin{overpic}[width=\figw\textwidth,trim={0cm 0cm 0cm 0cm},clip]{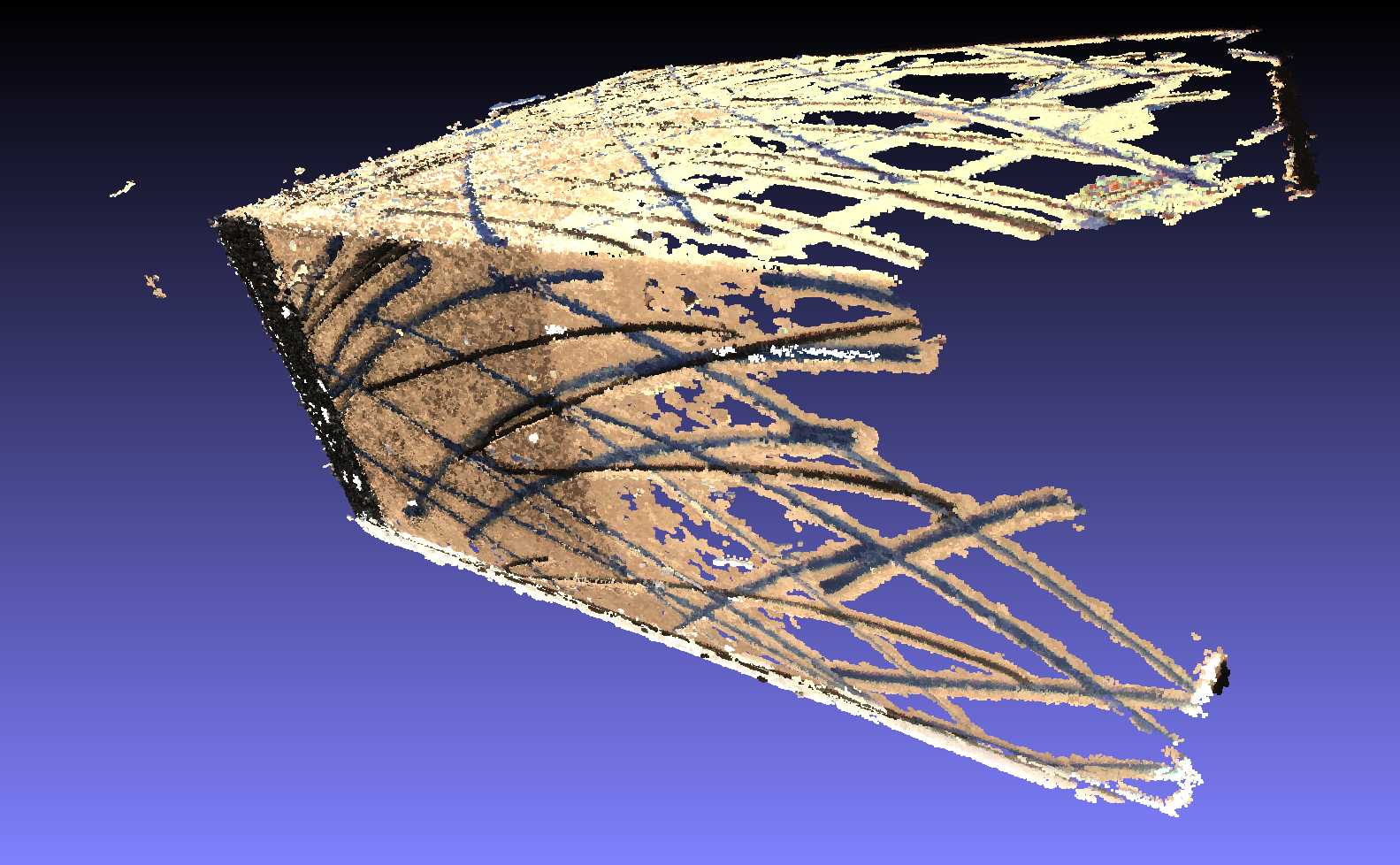}\end{overpic}
        \begin{overpic}[width=\figw\textwidth,trim={0cm 0cm 0cm 0cm},clip]{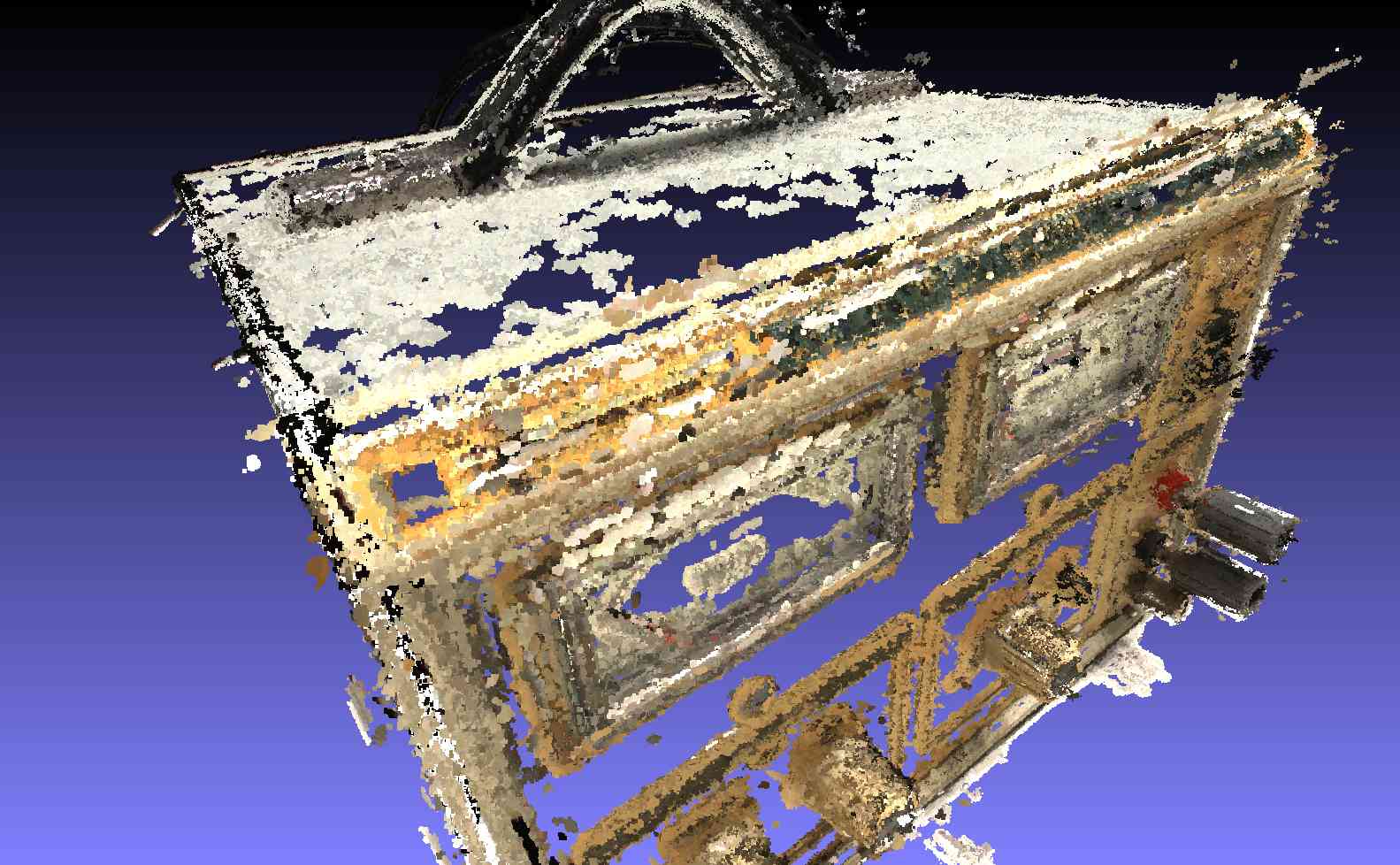}\end{overpic}
        \captionof{subfigure}{\textit{furu} \cite{furu2010accurate}}
    \end{minipage}
    \begin{minipage}[b]{\colw\textwidth}
        \begin{overpic}[width=\figw\textwidth,trim={0cm 0cm 0cm 0cm},clip]{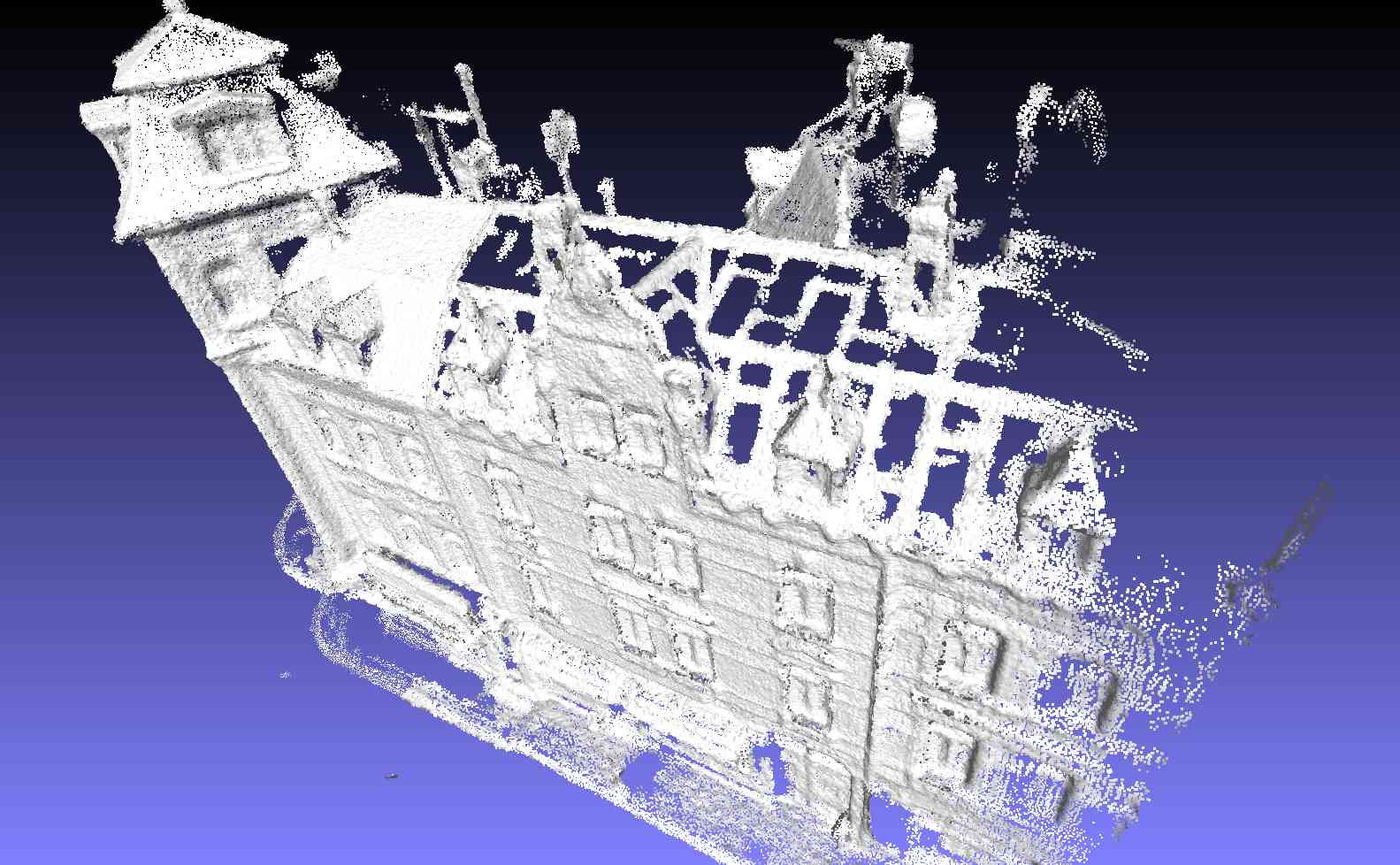}\end{overpic}
        \begin{overpic}[width=\figw\textwidth,trim={0cm 0cm 0cm 0cm},clip]{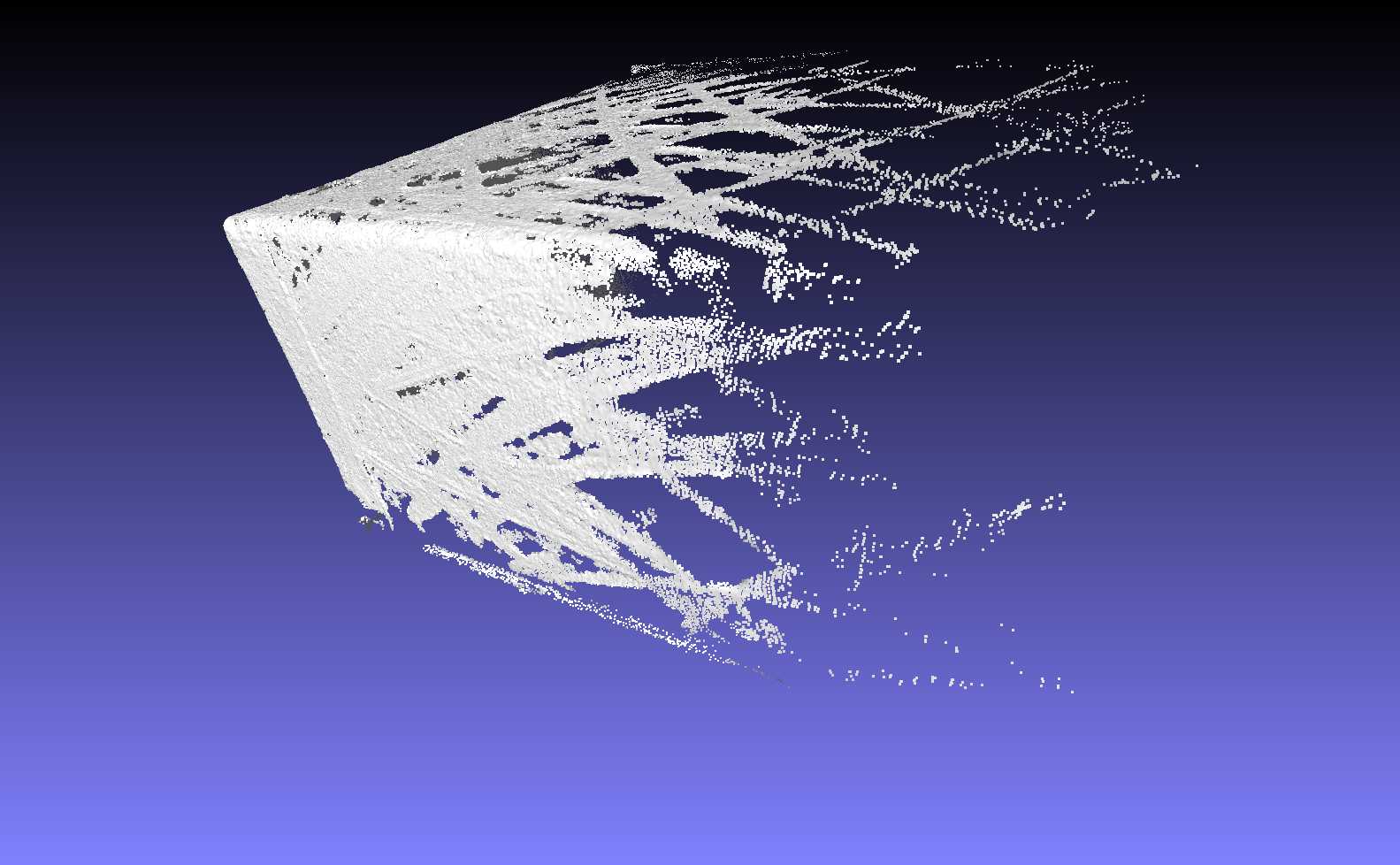}\end{overpic}
        \begin{overpic}[width=\figw\textwidth,trim={0cm 0cm 0cm 0cm},clip]{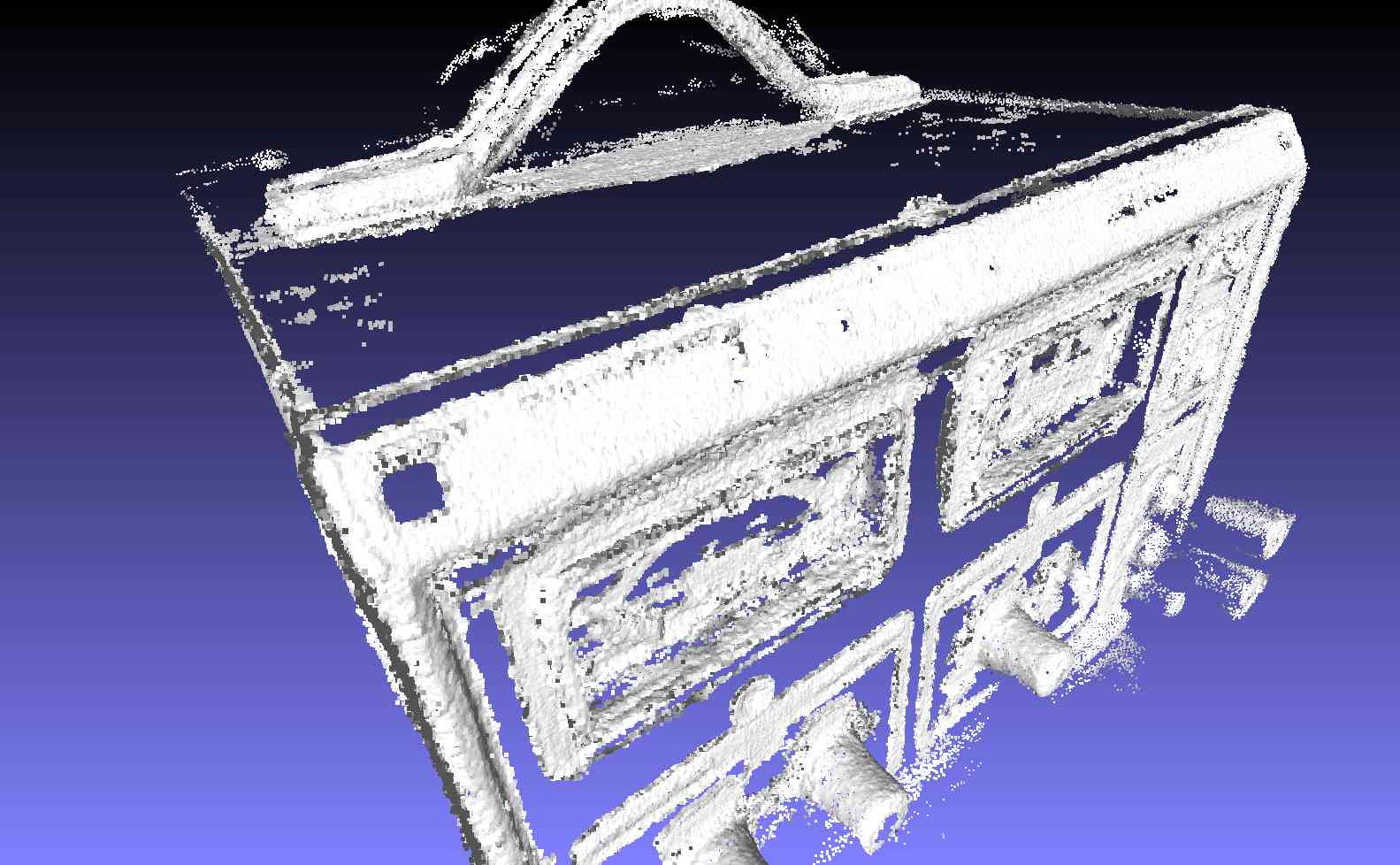}\end{overpic}
        \captionof{subfigure}{\textit{tola} \cite{tola2012efficient}}
    \end{minipage}
    \begin{minipage}[b]{\colw\textwidth}
        \begin{overpic}[width=\figw\textwidth,trim={0cm 0cm 0cm 0cm},clip]{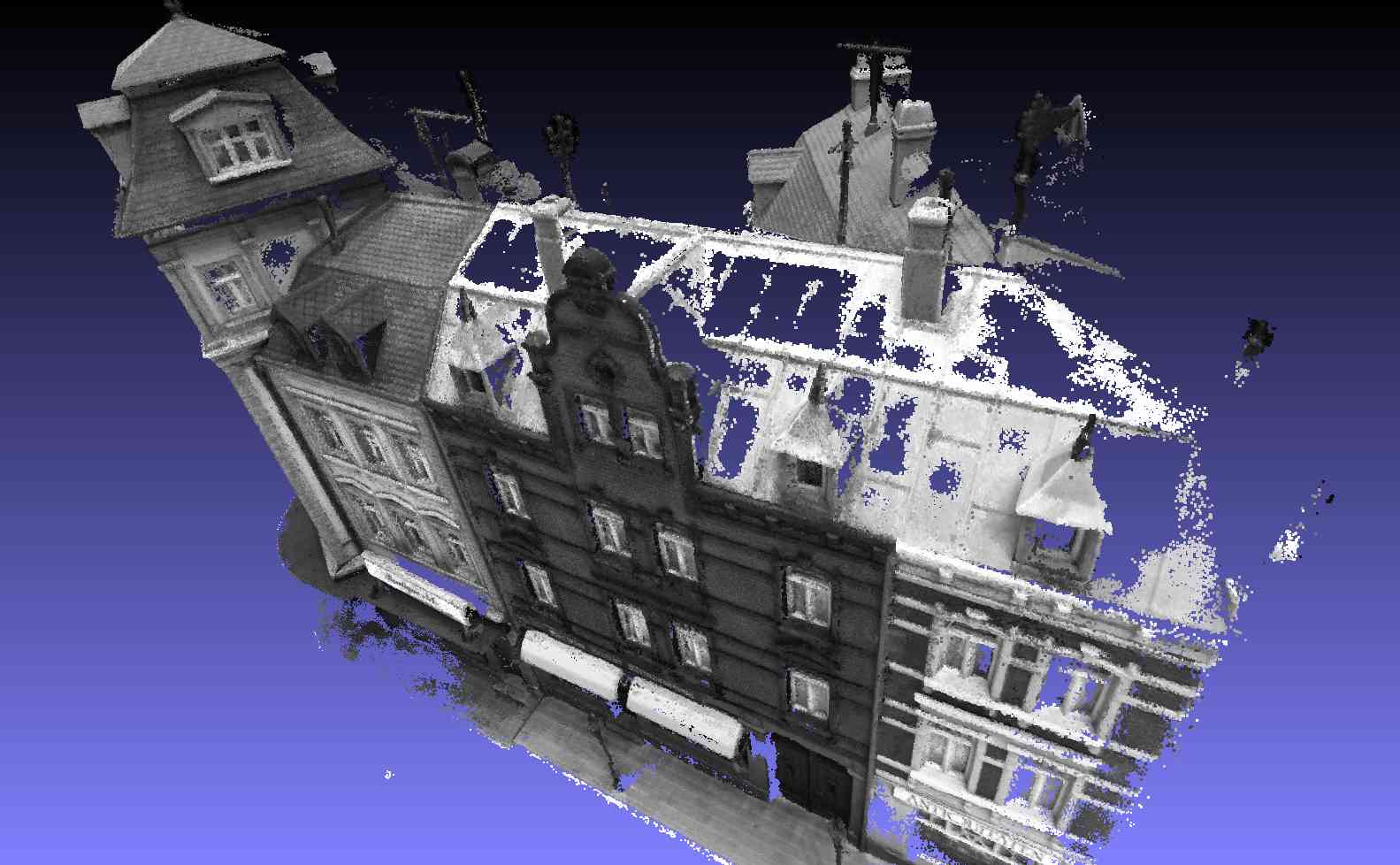}\end{overpic}
        \begin{overpic}[width=\figw\textwidth,trim={0cm 0cm 0cm 0cm},clip]{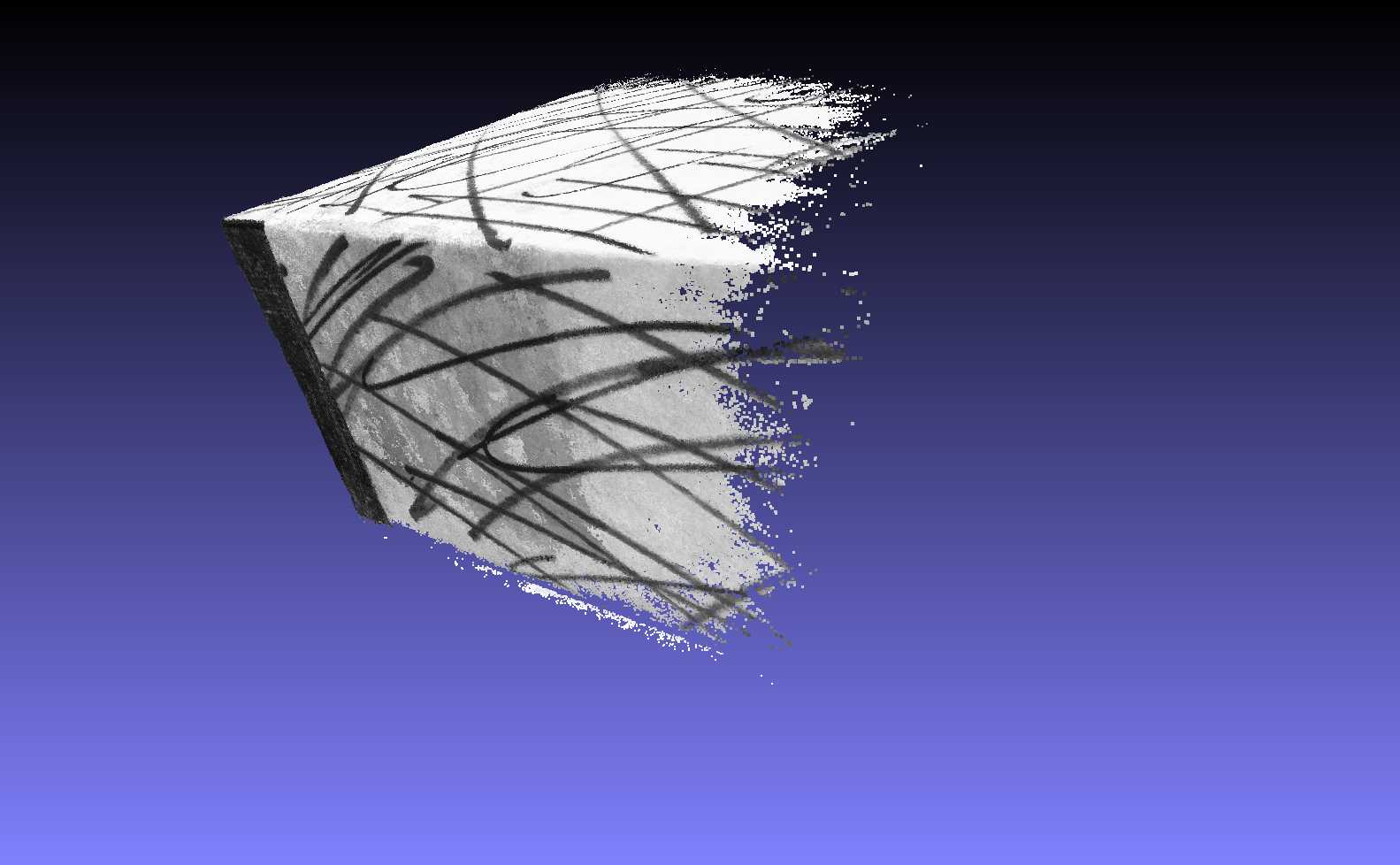}\end{overpic}
        \begin{overpic}[width=\figw\textwidth,trim={0cm 0cm 0cm 0cm},clip]{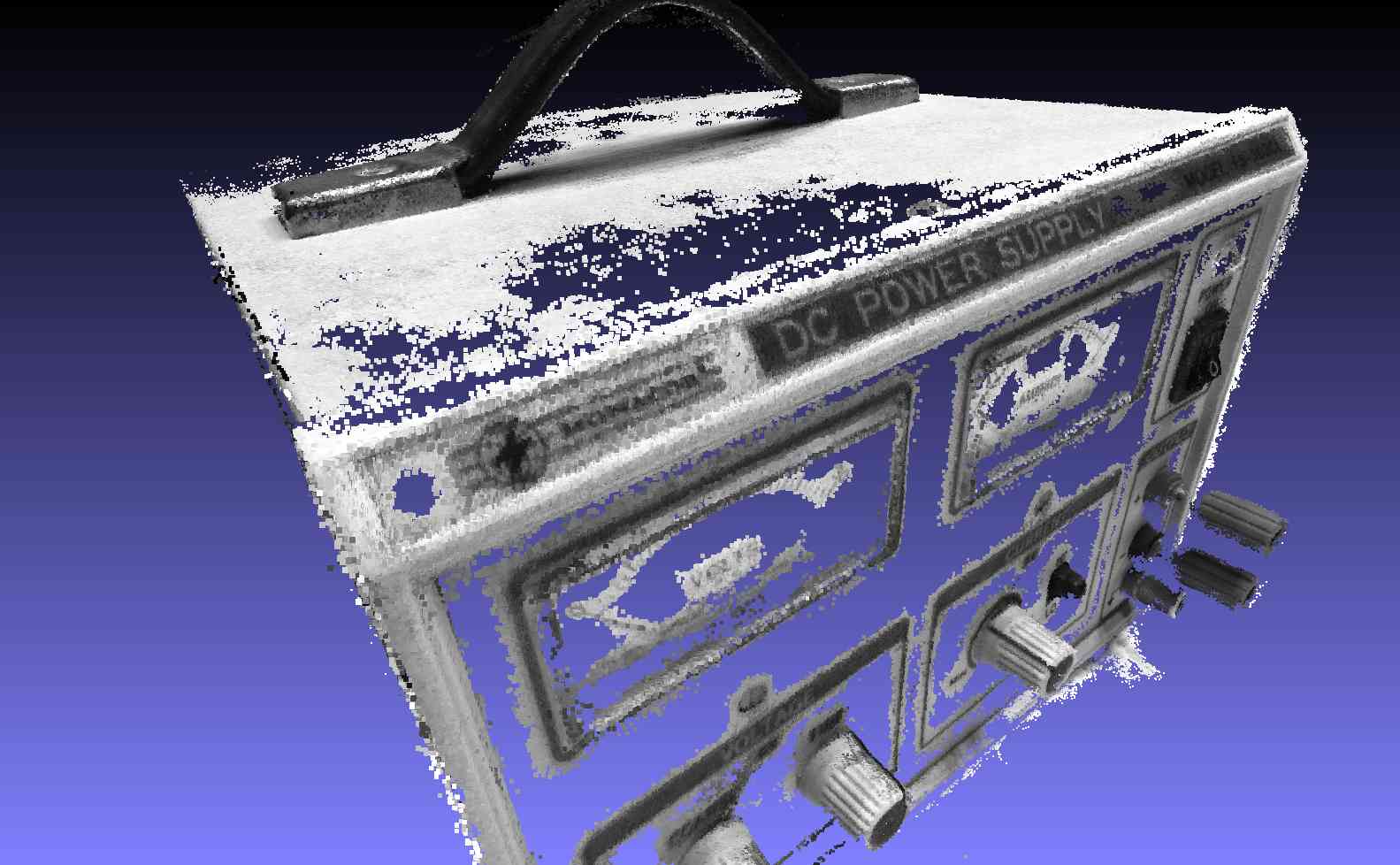}\end{overpic}
        \captionof{subfigure}{\textit{Gipuma} \cite{galliani2015massively}}
    \end{minipage}
    \caption{    
        Qualitative comparison to the reference model from the DTU dataset \cite{aanaes2016large} and the reconstructions obtained by  \cite{campbell2008using,furu2010accurate,tola2012efficient,galliani2015massively}. 
        The rows show the models 9, 10, 11 from the DTU dataset \cite{aanaes2016large}. 
    }
    \label{fig:methods}
\end{figure*}

\subsection{Comparison with others methods}

\begin{table}[t]
\footnotesize
\centering
\begin{tabular}{ccccc}
    methods (mm) & \begin{tabular}{@{}c@{}}mean \\ acc\end{tabular}
        & \begin{tabular}{@{}c@{}}med \\ acc\end{tabular}
        & \begin{tabular}{@{}c@{}}mean \\ compl\end{tabular}
        & \begin{tabular}{@{}c@{}}med \\ compl\end{tabular} \\
    \hline
    \textit{camp} \cite{campbell2008using}    &         0.834 &   0.335 &   \textbf{0.987} & \textbf{0.108} \\
    \textit{furu} \cite{furu2010accurate}    &          0.504 &   0.215 &   1.288 & 0.246 \\
    \textit{tola} \cite{tola2012efficient}    &             0.318 &   0.190 &   1.533 & 0.268 \\
    \textit{Gipuma} \cite{galliani2015massively}    &    \textbf{0.268} &   0.184 &   1.402 & 0.165 \\
    \hline
    \multicolumn{1}{l}{$s=32$, $\tau=0.7$, $\gamma=0\%$}    &    1.327 & 0.259 & 1.346 & 0.145 \\
    \multicolumn{1}{l}{$s=32$, $\tau=0.7$, $\gamma=80\%$}    &    0.779 & 0.204 & 1.407 & 0.172 \\
    \multicolumn{1}{l}{$s=32$,  adapt $\beta=6$,  $\gamma=80\%$} &    0.546 &   0.209 &   1.944 & 0.167 \\
    \hline
    \multicolumn{1}{l}{$s=64$, $\tau=0.7$, $\gamma=0\%$}    &    0.625 & 0.219 & 1.293 & 0.141 \\
    \multicolumn{1}{l}{$s=64$, $\tau=0.7$, $\gamma=80\%$}    &    0.454 & 0.191 & 1.354 & 0.164 \\
    \multicolumn{1}{l}{$s=64$,  adapt $\beta=6$,  $\gamma=80\%$} &    0.307 &   \textbf{0.183} &   2.650 & 0.342 \\
    \hline
\end{tabular}
\captionof{table}{
    Comparison with other methods. The results are reported for the test set consisting of 22 models.
} \label{tab:compare_methods}
\end{table}

We finally evaluate our approach on the test set consisting of 22 models and compare it with the methods \textit{camp} \cite{campbell2008using}, \textit{furu} \cite{furu2010accurate}, \textit{tola} \cite{tola2012efficient}, and \textit{Gipuma} \cite{galliani2015massively}. While we previously used cubes with $32^3$ voxels, we also include the results for cubes with $64^3$ voxels in Table~\ref{tab:compare_methods}. Using larger cubes, \ie $s=64$, improves accuracy and completeness using a constant threshold $\tau=0.7$ with $\gamma=0\%$ or $\gamma=80\%$. When adaptive thresholding is used, the accuracy is also improved but the completeness deteriorates.      

If we compare our approach using the setting $s=64$, $\tau=0.7$, and $\gamma=80\%$ with the other methods, we observe that our approach achieves a better accuracy than \textit{camp} \cite{campbell2008using} and \textit{furu} \cite{furu2010accurate} and a better completeness than \textit{tola} \cite{tola2012efficient} and \textit{Gipuma} \cite{galliani2015massively}. Overall, the reconstruction quality is comparable to the other methods. A qualitative comparison is shown in Fig.~\ref{fig:methods}.


\subsection{Runtime}

The training process takes about 5 days using an Nvidia Titan X GPU.
The inference is linear in the number of cubes and view pairs. 
For one view pair and a cube with $32^3$ voxels, inference takes 50ms. If the cube size is increased to $64^3$ voxels, the runtime increases to 400ms. However, the larger the cubes are the less cubes need to be processed. In case of $s=32$, model 9 is divided into 375k cubes, but most of them are rejected as described in Section~\ref{implementation} and only 95k cubes are processed. In case of $s=64$, model 9 is divided into 48k cubes and only 12k cubes are processed. 

\subsection{Generalization}

\begin{table}[htbp]
\centering
\footnotesize
\begin{tabular}{ccccc}
methods (mm) & \begin{tabular}{@{}c@{}}mean \\ acc\end{tabular}
        & \begin{tabular}{@{}c@{}}med \\ acc\end{tabular}
        & \begin{tabular}{@{}c@{}}mean \\ compl\end{tabular}
        & \begin{tabular}{@{}c@{}}med \\ compl\end{tabular} \\
    \hline
    49 views \\ \footnotesize$s=32$,  adapt $\beta=6$,  $\gamma=80\%$ &    0.197 &   0.149 &   1.762 & 0.237 \\
            \footnotesize$s=64$, $\tau=0.7$, $\gamma=80\%$    &    0.256 & 0.122 & \textbf{0.756} & \textbf{0.193} \\
    \hline
    15 views \\ \footnotesize$s=32$,  adapt $\beta=6$,  $\gamma=80\%$ &  \textbf{0.191}   &   0.135 &   2.517 & 0.387 \\
            \footnotesize$s=64$, $\tau=0.7$, $\gamma=80\%$    &    0.339 & \textbf{0.117} & 0.971 & 0.229 \\
    \hline
\end{tabular}
\captionof{table}{ 
Impact of the camera setting. The first two rows show the results for the 49 camera views, which are the same as in the training set. The last two rows show the results for 15 different camera views that are not part of the training set.   
The evaluation is performed for the three models 110, 114, and 118.
} \label{tab:15views}
\end{table}

In order to demonstrate the generalization abilities of the model, we apply it to a camera setting that is not part of the training set and an object from another dataset.  

The DTU dataset \cite{aanaes2016large} provides for some models additional 15 views which have a larger distance to the object. For training, we only use the 49 views, which are available for all objects, but for testing we compare the results if the approach is applied to the same 49 views used for training or to the 15 views that have not been used for training. In our test set, the additional 15 views are available for the models 110, 114, and 118. The results in Table \ref{tab:15views} show that the method performs very well even if the camera setting for training and testing differs. While the accuracy remains relatively stable, an increase of the completeness error is expected due to the reduced number of camera views.

\begin{figure}[htbp]
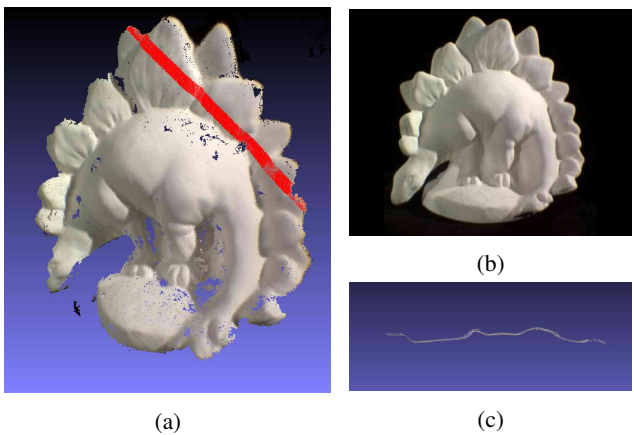

    \centering
    \begin{subfigure}[b]{0.52\linewidth}
        \begin{minipage}{\textwidth}
            \includegraphics[width=1\linewidth, height = 1\textheight, keepaspectratio=true,trim={9cm 0cm 11cm 0cm},clip]{{{figures/results/dinoSR/front01}}}  
            \caption{}
        \end{minipage}
    \end{subfigure}
    ~ 
    \begin{subfigure}[t]{0.45\linewidth}
        \begin{minipage}{\textwidth}
            \includegraphics[width=1\linewidth, height = 0.5\textheight, keepaspectratio=true,trim={0cm 4cm 0cm 0.5cm},clip]{{{figures/results/dinoSR/dinoSR0010-squ}}}   
            \caption{}
            \includegraphics[width=1\linewidth, height = 0.5\textheight, keepaspectratio=true,trim={8cm 8cm 8cm 8cm},clip]{{{figures/results/dinoSR/section01}}}
            \caption{}
        \end{minipage}%
    \end{subfigure}%
    \caption{
        (a) Reconstruction using only 6 images of the dinoSparseRing model in the Middlebury dataset \cite{seitz2006comparison}.
        (b) One of the 6 images.
        (c) Top view of the reconstructed surface along the red line in (a).
    }
   \label{fig:middlebury}
\end{figure}


We also applied our model to an object of the Middlebury MVS dataset \cite{seitz2006comparison}. We use
6 input images from the view ring of the model dinoSparseRing for reconstruction. The reconstruction is shown in Fig.~ \ref{fig:middlebury}.

\section{Conclusion}

In this work, we have presented the first end-to-end learning framework for multiview stereopsis. To efficiently encode the camera parameters, we have also introduced a novel representation for each available viewpoint. The so-called colored voxel cubes combine the image and camera information and can be processed by standard 3D convolutions. Our qualitative and quantitative evaluation on a large-scale MVS benchmark demonstrated that our network can accurately reconstruct the surface of 3D objects. While our method is currently comparable to the state-of-the-art, the accuracy can be further improved by more advanced post-processing methods. We think that the proposed network can also be used for a variety of other 3D applications.

\section*{Acknowledgments} 

The work has been supported in part by Natural Science Foundation of China (NSFC) under contract
No. 61331015, 6152211, 61571259 and 61531014, in part by the National key foundation
for exploring scientific instrument No.2013YQ140517, in part by Shenzhen Fundamental
Research fund (JCYJ20170307153051701) and in part by the DFG project GA 1927/2-2 as part of the DFG Research Unit FOR 1505 Mapping on Demand (MoD).



{\small
\bibliographystyle{ieee}
\bibliography{egbib}
}

\end{document}